\documentclass[sigconf]{acmart}

\AtBeginDocument{%
  }

\acmDOI{10.1145/3746027.3755294}

\acmConference[MM '25]{Proceedings of the 33rd ACM International Conference on Multimedia}{October 27--31, 2025}{Dublin, Ireland}

\acmBooktitle{Proceedings of the 33rd ACM International Conference on Multimedia (MM '25), October 27--31, 2025, Dublin, Ireland}
\acmISBN{979-8-4007-2035-2/2025/10}
\usepackage{enumitem}
\usepackage{xcolor}
\usepackage{colortbl}
\newcommand{\MYhref}[3][blue]{\href{#2}{\color{#1}{#3}}} 
\definecolor{mred}{RGB}{238, 34, 12}
\definecolor{mgreen}{RGB}{1, 127, 0}
\definecolor{mblue}{RGB}{0, 77, 158}
\newcommand{\mredbf}[1]{\textcolor{mred}{\textbf{#1}}}
\newcommand{\mbluebf}[1]{\textcolor{mblue}{\textbf{#1}}}

\definecolor{darkgreen}{RGB}{0, 100, 0} 
\usepackage{arydshln} 
\usepackage{pifont}
\usepackage{wasysym}

\usepackage[skip=3pt]{caption}
\usepackage{titlesec}
\begin{document}

\title{LMM4Edit: Benchmarking and Evaluating Multimodal \\Image Editing with LMMs}

\author{Zitong Xu\textsuperscript{1}, Huiyu Duan\textsuperscript{1}\textsuperscript{†}, Bingnan Liu\textsuperscript{2}, Guangji Ma\textsuperscript{2}, Jiarui Wang\textsuperscript{1}, Liu Yang\textsuperscript{1},\\
Shiqi Gao\textsuperscript{1}, 
Xiaoyu Wang\textsuperscript{2}, Jia Wang\textsuperscript{1}, Xiongkuo Min\textsuperscript{1}\textsuperscript{†}\, Guangtao Zhai\textsuperscript{1}\textsuperscript{†}, Weisi Lin\textsuperscript{3}}
\author{\textsuperscript{1}Institute of Image Communication and Network Engineering, Shanghai JiaoTong University\\
\textsuperscript{2}University of Electronic and Science Technology of China, \textsuperscript{3}Nanyang Technological University \\
$\{$xuzitong, huiyuduan, guangjima wangjiarui, ylyl.yl, shiqigao, jiawang, minxiongkuo, zhaiguangtao$\}$@sjtu.edu.cn, wslin@ntu.edu.sg}
\authornote{† Corresponding authors.\\
This work was supported in part by the National Natural Science Foundation of China under Grants 62401365, 62225112, 62271312, 62132006, U24A20220, and in part by the China Postdoctoral ScienceFoundation under Grant Number BX20250411, 2025M773473.}

\renewcommand{\shortauthors}{}

\begin{abstract}
The rapid advancement of Text-guided Image Editing (TIE) enables image modifications through text prompts. However, current TIE models still struggle to balance image quality, editing alignment, and consistency with the original image, limiting their practical applications. Existing TIE evaluation benchmarks and metrics have limitations on scale or alignment with human perception. To this end, we introduce \textbf{EBench-18K}, the first large-scale image \underline{E}diting \underline{Bench}mark including \underline{18K} edited images with fine-grained human preference annotations for evaluating TIE. Specifically, EBench-18K includes 1,080 source images with corresponding editing prompts across 21 tasks, 18K+ edited images produced by 17 state-of-the-art TIE models, 55K+ mean opinion scores (MOSs) assessed from three evaluation dimensions, and 18K+ question-answering (QA) pairs. Based on EBench-18K, we employ outstanding LMMs to assess edited images, while the evaluation results, in turn, provide insights into assessing the alignment between the LMMs' understanding ability and human preferences. Then, we propose \textbf{LMM4Edit}, a \underline{LMM}-based metric for evaluating image \underline{Edit}ing models from perceptual quality, editing alignment, attribute preservation, and task-specific QA accuracy in an \textit{all-in-one} manner. Extensive experiments show that LMM4Edit achieves outstanding performance and aligns well with human preference. Zero-shot validation on the other datasets also shows the generalization ability of our model. The dataset and code are available at \MYhref[magenta]{https://github.com/IntMeGroup/LMM4Edit}{https://github.com/IntMeGroup/LMM4Edit}.
\end{abstract}
\vspace{-10em}

\vspace{-4em}
\begin{teaserfigure}
\centering
  \includegraphics[width=\textwidth]{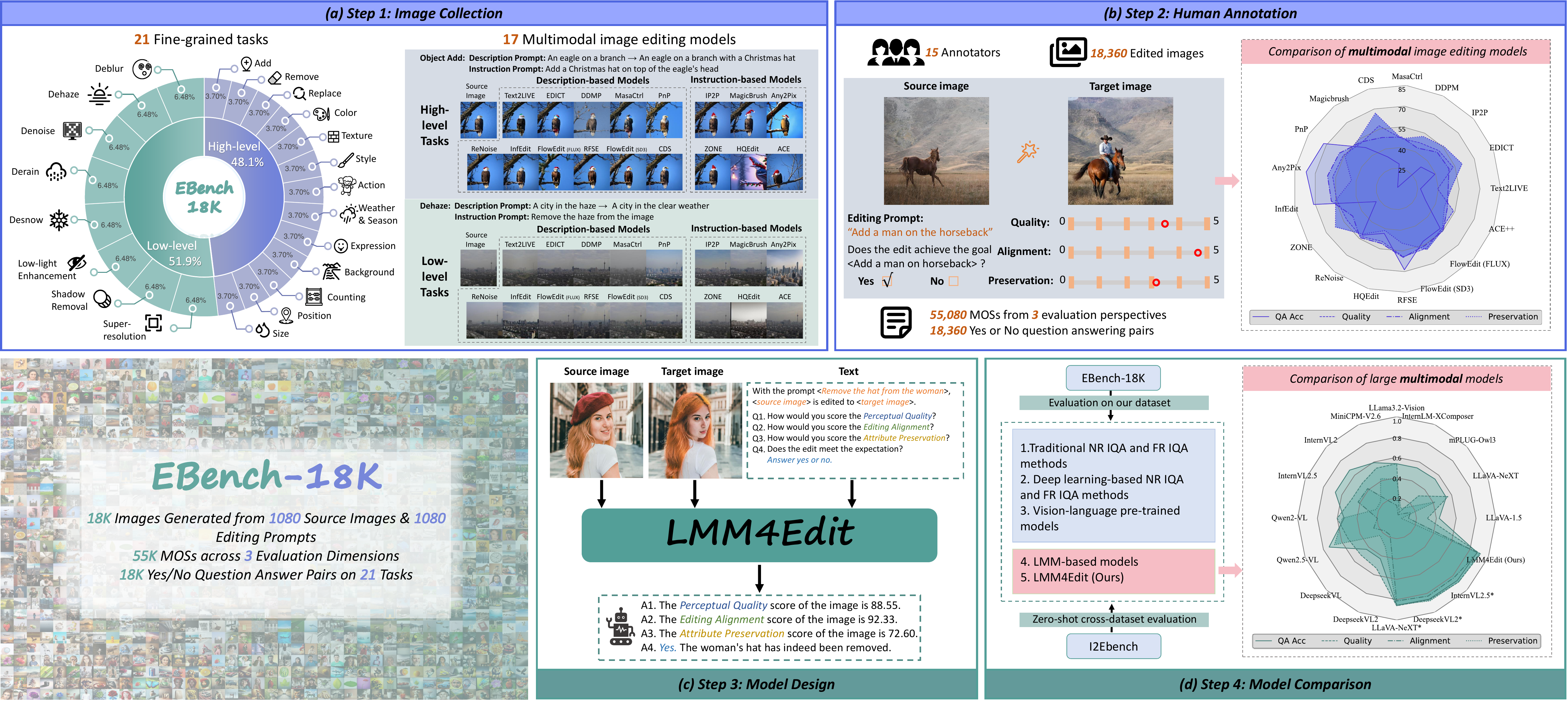}
  \caption{We present the text-guided image editing (TIE) evaluation database and model, termed EBench-18K and LMM4Edit, respectively. (a) We first collect 1080 source images and 1080 comprehensive editing prompts across 21 fine-grained tasks. Then 17 TIE models are applied to generate 18K images. (b) 55K MOSs and 18K question-answering pairs are acquired from 15 annotators. (c) We design LMM4Edit to evaluate TIE models. (d) We perform model comparisons on EBench-18K and conduct a zero-shot cross-dataset evaluation.}
  \label{overall}
\end{teaserfigure}

\maketitle
\setlength{\textfloatsep}{1.5pt plus 0.2pt minus 0.5pt}
\setlength{\dbltextfloatsep}{1.5pt plus 0.2pt minus 0.5pt}
\setlength{\dblfloatsep}{1.5pt plus 0.2pt minus 0.5pt}
\setlength{\intextsep}{1.5pt plus 0.2pt minus 0.5pt}
\setlength{\abovecaptionskip}{1.5pt plus 0.2pt minus 0.5pt}
\setlength{\belowcaptionskip}{1.5pt plus 0.2pt minus 0.5pt}
\setlength{\abovedisplayskip}{2pt}
\setlength{\belowdisplayskip}{2pt}
\setlength{\topsep}{1.5pt}
\titlespacing*{\section}{0pt}{1.5pt plus 0.2pt minus 0.5pt}{1.5pt plus 0.2pt minus 0.5pt} 
\titlespacing*{\subsection}{0pt}{1.2pt plus 0.2pt minus 0.5pt}{1.2pt plus 0.2pt minus 0.5pt} 

\begin{table*}[t]
\fontsize{7.5}{8}\selectfont
\setlength{\arrayrulewidth}{0.275pt}  
\setlength{\heavyrulewidth}{0.08em}  
\centering
\caption{Comparison of text-guided image editing model evaluation benchmarks. \textcolor{darkgreen}{\ding{51}} means publicly available, \textcolor{red}{\ding{55}} means not applicable or not available.}
 \resizebox{1\textwidth}{!}{
\begin{tabular}{lcccccccccc}
\toprule
\noalign{\vspace{-1.5pt}}
Databases& MOSs  & Source Images &Edited Images & Annotations& Models & Tasks &Prompt Type  & Dimensions & QA Pairs \\
\hline
\noalign{\vspace{1.5pt}}
TedBench \cite{TedBench}&\color{red}{\ding{55}}    & 100&100& \color{red}{\ding{55}} &3 & 4&Description & Human Preference & \color{red}{\ding{55}} \\
EditVal \cite{EditVal} & \color{red}{\ding{55}}   &648&\color{red}{\ding{55}}& \color{red}{\ding{55}} &8&12&Instruction&Alignment, Preservation&\color{red}{\ding{55}}\\
EditEval \cite{ESurvey} & \color{red}{\ding{55}} & 150&\color{red}{\ding{55}}&\color{red}{\ding{55}} & 19& 7&Description, Instruction &Quality, Alignment, Preservation, Realism&\color{red}{\ding{55}}\\
I2EBench \cite{I2EBench}&\color{red}{\ding{55}}     &2,218&17,744&17,744&8&16&Instruction&Alignment&\textcolor{darkgreen}{\ding{51}}\\
EmuEdit \cite{EmuEdit} &\color{red}{\ding{55}}     &3,055& \color{red}{\ding{55}}& \color{red}{\ding{55}}&5&9&Description, Instruction&Alignment, Preservation&\color{red}{\ding{55}} \\
\hline
\noalign{\vspace{1.5pt}}
  \rowcolor{gray!20}  
 \textbf{EBench-18K} & \textcolor{darkgreen}{\ding{51}} &  \textbf{1,080}&\textbf{18,360}& \textbf{
 1,101,600}& \textbf{17}& \textbf{21} & \textbf{Description, Instruction}& \textbf{Quality, Alignment, Preservation}&\textcolor{darkgreen}{\ding{51}} \\
 \noalign{\vspace{-1.5pt}}
\bottomrule
\label{comparison}
\end{tabular}}
\end{table*}
\section{Introduction}
The rapid advancement of Text-guided Image Editing (TIE) allows for image modifications through text prompts  \cite{ESurvey, ACE, Flowedit, ip2p, Magicbrush}. However, state-of-the-art TIE models still struggle to balance perceptual quality, editing alignment, and attribute preservation, limiting their reliability and practicality in real-world applications  \cite{IEBench}. Given that human evaluation is costly and inefficient, it is crucial to develop effective automatic evaluation metrics that closely align with human perception and preferences.

Existing TIE evaluation methods include image quality assessment (IQA) metrics  \cite{NQM,MSSIM,SSIM,FSIM,IFC,VIF,BIQI,BRISQUE,LPIPS}, vision-language approaches \cite{imagereward,clipscore,pickscore} and LMM-based evaluation methods \cite{ESurvey,I2EBench}. Traditional IQA metrics primarily assess natural distortions such as noise, blur, compression, \textit{etc.}, but they fail to capture key challenges in TIE, such as structural distortions, text-image misalignment, and discrepancies between the source and target images. While vision-language approaches have made significant progress in text-to-image generation evaluation by incorporating human visual feedback \cite{IQAinstruction,aigciqa2023}, they focus solely on alignment between text and image, neglecting the editing alignment and relationship between the source and edited images. Recent studies have explored using LMMs for general quality evaluation \cite{aigv,finevq}, and some works have employed LMMs to bench TIE models \cite{ESurvey, I2EBench}. However, these zero-shot results fail to align with human preferences in all dimensions. Additionally, existing TIE evaluation benchmarks assess only a limited set of TIE models or only consider the alignment dimension, limiting their generalization and practical applicability.

In this paper, we introduce \textbf{EBench-18K}, a large-scale image \underline{E}diting \underline{Bench}mark to evaluate human preferences for TIE, as shown in Figure~\ref{overall}(a)(b). The dataset includes \textbf{1,080} high-quality source images from the free photography website and open datasets, accompanied by corresponding diverse editing prompts across \textbf{21 editing tasks}. Based on these source images and editing prompts, we generate \textbf{18K+} edited images using \textbf{17 state-of-the-art TIE models}. Through an extensive subjective study, we collect \textbf{1M+} human annotations evaluated from perceptual quality, editing alignment, attribute preservation, and task-specific accuracy, respectively, which result in 55,080 high-quality \textbf{mean opinion scores (MOSs)} and 18,360 question-answer (QA) pairs for TIE evaluation.

Based on EBench-18K, we propose \textbf{LMM4Edit}, a novel an \underline{LMM}-based all-in-one approach for evaluating image \underline{Edit}ing models from multiple dimensions, including perceptual quality, editing alignment, attribute preservation, and task-specific accuracy, as shown in Figure~\ref{overall}(c). Specifically, LMM4Edit is built upon a LMM backbone fine-tuned with instruction tuning  \cite{Ituning}. To enhance the performance, we apply adaptive low-rank adaptation (AdaLoRA)  \cite{adalora} to both the vision encoder and the language model, refining their ability to capture quality-aware, instruction-relevant and preservation-oriented attributes. A two-stage training step is used to achieve the better score regression. Extensive experiments on EBench-18K demonstrate that LMM4Edit achieves state-of-the-art performance and good generalization ability. 
\begin{table}[t]
\centering

\caption{An overview of text-guided image editing methods selected to construct our EBench-18K.}
\fontsize{4}{4.5}\selectfont
\setlength{\arrayrulewidth}{0.15pt}  
\setlength{\heavyrulewidth}{0.08em}  
\setlength{\cmidrulewidth}{0.15pt}    
 \resizebox{0.48\textwidth}{!}{\begin{tabular}{lcccc}
\toprule
 \noalign{\vspace{-1.5pt}}

Models & Time &Prompt Type  & Method &Resolution\\

\hline
 \noalign{\vspace{1.5pt}}
Text2LIVE  \cite{Text2live} & 2022.04 &Description  &GAN \cite{GAN} &512$\times$512\\
EDICT  \cite{EDICT} & 2022.11 &Description  & SD1.4 \cite{SD} &512$\times$512\\
IP2P \cite{ip2p}& 2023.04 &Instruction  &SD1.4 \cite{SD} &768$\times$768\\
DDPM \cite{DDPM} & 2023.04 &Description  & SD2.1  \cite{SD}&512$\times$512\\
MasaCtrl \cite{Masactrl} & 2023.04 &Description  &SD1.4  \cite{SD}&512$\times$512\\
CDS \cite{CDS} & 2023.11 &Description  & SD1.4  \cite{SD}&512$\times$512\\
Magicbrush \cite{Magicbrush} & 2023.06 &Instruction  & SD1.4  \cite{SD}&768$\times$768\\
PnP \cite{PnP} & 2023.10 &Description  & SD1.5 \cite{SD} &512$\times$512\\
Any2Pix \cite{instructany2pix} & 2023.12 &Instruction  & SDXL \cite{sdxl} &1024$\times$1024\\
InfEdit \cite{InfEdit} & 2023.12 &Description  & SD1.4 \cite{SD} &512$\times$512\\
ZONE \cite{zone} & 2023.12 &Instruction  & SD1.5 \cite{SD} &512$\times$512\\
ReNoise \cite{renoise} & 2024.03 &Description  & SDXL \cite{sdxl} &512$\times$512\\
HQEdit \cite{HQ} & 2024.04 &Instruction  & DIFT \cite{DIFT} &512$\times$512\\
RFSE \cite{RFSE} & 2024.11 &Description  & FLUX \cite{FLUX} &1024$\times$768\\
FlowEdit (SD3)  \cite{Flowedit} & 2024.12 &Description  & SD3 \cite{SD} &1024$\times$1024\\
FlowEdit (FLUX)  \cite{Flowedit} & 2024.12 &Description & FLUX \cite{FLUX} &1024$\times$1024\\
ACE++  \cite{ACE} & 2025.01 &Instruction  & FLUX \cite{FLUX}  &1024$\times$1024\\
 \noalign{\vspace{-1.5pt}}
\bottomrule
\end{tabular}}
\label{TIE models}
\end{table}
The main contributions of this work include:
\begin{itemize}[left=0pt, labelsep=0.6em, labelwidth=0pt]
    \item We introduce EBench-18K, a large-scale dataset containing 18K edited images across diverse tasks with over 1M+ human annotations covering perceptual quality, editing alignment, attribute preservation and task-specific accuracy dimensions.
    \item We use EBench-18K to bench both TIE generation ability and the LMMs' understanding and evaluating capabilities.
    \item We propose LMM4Edit, a novel LMM-based all-in-one metric providing fine-grained perceptual quality, editing alignment, attribute preservation assessments for TIE models.
    \item Extensive experimental results on EBench-18K validate the superior performance of LMM4Edit and its strong in aligning with human perception and generalization ability.
\end{itemize}

\begin{figure*}  \includegraphics[width=\textwidth,height=0.13\textwidth]{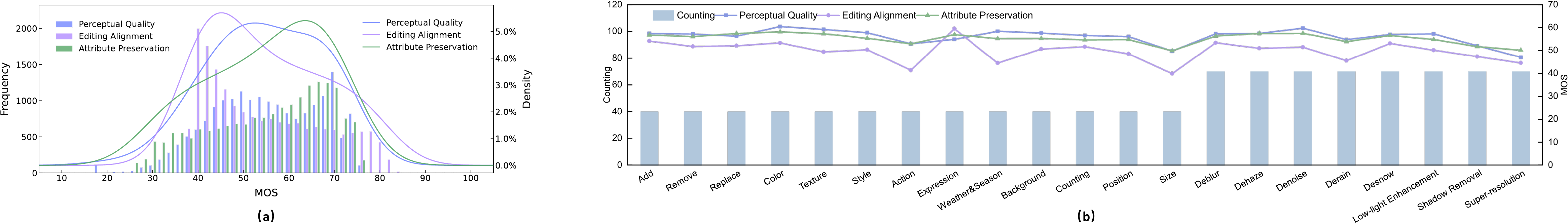}
  \caption{(a) Distribution of perceptual quality, editing alignment and attribute preservation MOSs. (b) Counts and average MOSs across different tasks.}
  \label{dataanalysis}
\end{figure*}
\begin{figure*}
  \includegraphics[width=\textwidth,height=0.16\textwidth]{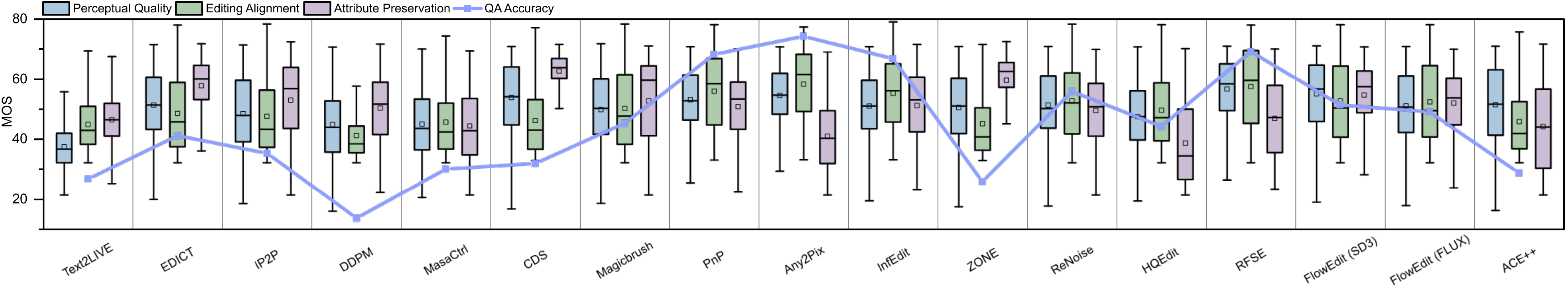}
  \caption{Comparison of TIE models across the dimensions of perceptual quality, editing alignment, attribute preservation MOSs, and question-answer (QA) accuracy.}
  \label{allcomparison}
\end{figure*}
\section{Related Work}
\subsection{Text-guided Image Editing}
With the advancement of generative models such as Stable Diffusion  \cite{SD} and FLUX  \cite{FLUX}, numerous TIE methods have emerged \cite{ESurvey,I2EBench}. TIE can be categorized into description-based approaches, which modify images based on text descriptions before and after editing (\textit{e.g.,} "A little dog wearing glasses" $\rightarrow$ "A little dog") and instruction-based approaches, which directly follow the editing commands (\textit{e.g.,} "Remove the glasses")  \cite{ESurvey}. For description-based methods, some approaches, such as  \cite{DDS, SLS, PDS}, utilize an optimization process that adjusts the image to better align with the user-provided prompt. In contrast, zero-shot editing methods like  \cite{DDPM, Masactrl, CDS, PnP, InfEdit} avoid the need for specific optimization, which first use image-to-noise inversion techniques to obtain the latent representation, then generate the edited image based on new prompts. For instruction-based editing methods, Instructpix2pix \cite{ip2p} introduces instruction-based editing using a large-scale pair-wise editing dataset, later improved by Magicbrush \cite{Magicbrush} with manually annotated data. Recent advances further integrate large language models (LLMs) to enhance the editing-instruction understanding  \cite{HQ, ACE}.
\subsection{Benchmarks for Text-guided Image Editing}
\setlength{\tabcolsep}{3pt}

There are few benchmarks for TIE, and the limitations of them and the comparisons between these benchmarks and our EBench-18K are shown in Table~\ref{comparison}. TedBench  \cite{TedBench}, the first TIE benchmark, provides a very small dataset with only 100 source images with prompts, and 100 edited images. EditVal  \cite{EditVal} includes broader tasks and methods, but its images suffer from low resolution and blurriness due to the method limitation. EditEval \cite{ESurvey} includes numerous TIE models, however, the scores directly derived from LMMs do not align well with human perception, as shown in our experiments. I2EBench  \cite{I2EBench} offers a variety of tasks, but lacks description-based TIE models. EmuEdit \cite{EmuEdit} includes comprehensive instruction and description prompts for various editing tasks, however, it lacks an automatic evaluation metric.
Moreover, none of the above existing benchmarks incorporate MOSs, making it difficult to comprehensively measure the alignment between the metrics and human preferences. Our EBench-18K stands out by providing high-quality edited images produced by diverse models and tasks, with corresponding perceptual quality, editing alignment, attribute preservation MOSs, and task-specific QA pairs.
\subsection{Metrics for Text-guided Image Editing}
Numerous image quality assessment (IQA) models have been proposed, including full-reference (FR) IQA \cite{SSIM, FSIM, GMSD, LPIPS, STLPIPS, AHIQ} and no-reference (NR) IQA \cite{BRISQUE, BLII, NIQE, CNN, Hyper, MANIQA} metrics. 
NR IQA methods relie no reference image for comparison, including handcrafted approaches (\textit{e.g.,} BRISQUE, BLIINDS-II, NIQE) and deep learning-based methods (\textit{e.g.,} CNNIQA, HyperIQA, MANIQA). While these models capture quality-aware features to predict IQA scores, they fail to evaluate the editing alignment and source-target relation. FR IQA methods include handcrafted approaches (\textit{e.g.,} SSIM, FSIM, GMSD) and deep learning-based methods (\textit{e.g.,} LPIPS, ST-LPIPS, AHIQ). 
Though the popular FR metrics can measure the differences between source and target images, they cannot understand and use the editing attributes for evaluation. Some vision-language metrics such as ImageReward \cite{imagereward}, CLIPScore \cite{clipscore}, and PickScore \cite{pickscore} have been proposed to evaluate the alignment for text-based image generation. However, they may fail to effectively evaluate editing alignment and source-target relationships. With the advancement of LMMs, some LMMs demonstrates effectiveness in describing image quality and performing question-answering tasks \cite{LMMIQA}. These models also support multiple image inputs, making them suitable for TIE evaluation. Since the multi-dimensional alignment between LMMs and human preferences for evaluating TIE is still underexplored, in this paper, we use EBench-18K to bench both TIE models and LMM models, and propose LMM4Edit to better evaluate TIE.
\begin{figure*}
  \includegraphics[width=0.95\textwidth,height=0.19\textwidth]{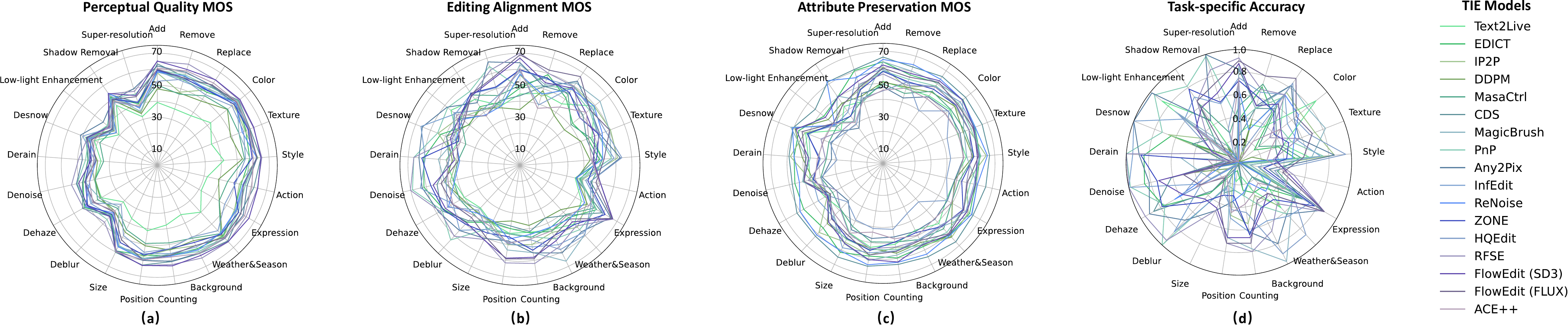}
  \caption{Comparison of human evaluations on 17 TIE models across 21 tasks in terms of perceptual quality, editing alignment, attribute preservation, and task-specific accuracy, respectively.}
  \label{MOScomparison}
\end{figure*}
\section{EBench-18K}
In this section, we introduce our \textbf{EBench-18K}, the first large-scale TIE dataset with fine-grained scores. This database comprises 1080 high-quality source images with both instruction prompts and description prompts on 21 tasks, 18,360 edited images from 17 TIE models, and 1,101,600 human annotations on perceptual quality, editing alignment, attribute preservation and task-specific QA pairs. With its broad range of image content, EBench-18K provides a comprehensive resource for TIE evaluation and can also serve as a tool for assessing the interpretation ability of LMMs.
\subsection{Design of Editing Tasks} 
Taking both practical applications and widespread usage into account, we first select 21 TIE tasks, categorized into \textbf{high-level tasks} and \textbf{low-level tasks}. High-level tasks evaluate model’s ability to accurately interpret editing prompts and apply precise modifications to specific areas of the input images, while low-level tasks focus on the global adjustment and fine-grained image processing.

As shown in Figure~\ref{overall}(a), the high-level tasks encompass 13 dimensions: add, remove, replace, color, texture, style, action, expression, weather\&season, background, counting, position, and size. The low-level tasks cover 8 dimensions: deblur, dehaze, denoise, derain, desnow, low-light enhancement, shadow removal, and super-resolution.
\subsection{Data Collection}
For each high-level task, we collect 40 distinct source images from free photography websites, ensuring a minimum resolution of 1024×1024, as this matches or exceeds the maximum resolution of images required by TIE models. For each low-level task, we select 70 source images from relevant datasets. Specifically, the low-level task set contains 8 subsets, including the Deblur subset from GoPro \cite{GoPro}, the Dehaze subset from exBeDDE \cite{exBeDDE}, the Denoise subset from SSID \cite{SSID}, the Derain subset from RainyDataset \cite{derain}, the Desnow subset from CSD \cite{CSD}, the Low-light Enhancement subset from LOL \cite{LOL}, the Shadow Removal subset from ISTD \cite{ISTD}, and the Super-Resolution subset from various online sources, with images downsampled to lower resolutions. In total, we collect 1,080 source images (40 × 13 high-level tasks + 70 × 8 low-level tasks).

For each source image, we manually design an initial instruction prompt based on its content. Within each task, instructions vary, with approximately 10$\%$ complex editing requirements. We then use the advanced LMM, InternVL2.5 \cite{internViT}, to generate description prompts (source description prompts and target description prompts) based on source images and instruction prompts, and manually screen and correct these prompts. This ensures clarity and detail, allowing description-based TIE models to generate images that better align with editing expectations. Totally, we collect 3240 prompts (1080 instruction prompts + 1080 source description prompts + 1080 target description prompts).

Next, we select 17 state-of-the-art TIE models, covering both description-based and instruction-based approaches, as listed in Table~\ref{TIE models}. Finally, using our source images and editing prompts, we generate a total of 18,360 edited images (1,080 source images × 17 TIE models).

\subsection{Subjective Experiment Setup}
To evaluate the edited images, we conduct a subjective quality assessment experiment using the EBench-18K database. This experiment is designed to capture human preferences for edited images, ensuring the results align with real-world human perception.

As shown in Figure~\ref{overall}(b), during the experiment, participants are presented with a source image, an edited image, and an editing prompt for each evaluation. Participants are asked to assess the edited image using a 5-point continuous scale from three aspects:
\begin{itemize}[left=0pt, labelsep=0.6em, labelwidth=0pt]
    \item \textbf{Perceptual Quality:} focuses on assessing the overall quality of the edited images, considering factors such as the authenticity, distortion, color accuracy, and detail richness. 
   \item \textbf{Editing Alignment:} evaluates how accurately the edited images align with the editing instructions, \textit{i.e.}, assesses the precision of the modifications.
   \item \textbf{Attribute Preservation:} assesses how well the edited image preserves the context of the source image in regions where the change is not expected.  
\end{itemize}  
Moreover, in addition to the ratings, participants are instructed to answer a \textbf{task-specific yes/no question} for each image to determine whether generated image meet the editing expectation.

The experiment is conducted using a Python-based GUI displayed on a calibrated LED monitor with a resolution of 3840 $\times$ 2160, with images shown in 512$\times$512 resolution in a random order. A total of 30 professional annotators, seated 2 feet from the monitor in a controlled environment, complete the study in 45 sessions, each under 30 minutes, to mitigate fatigue. Each image is assessed by 15 participants. In total, we collect 826,200 score ratings (15 annotators $\times$ 18,360 edited images $\times$ 3 dimensions) and 275,400 question-answer pairs (15 annotators $\times$ 18,360 edited images).

\begin{figure*}[t]
  \includegraphics[width=0.9\textwidth,height=0.28\textwidth]{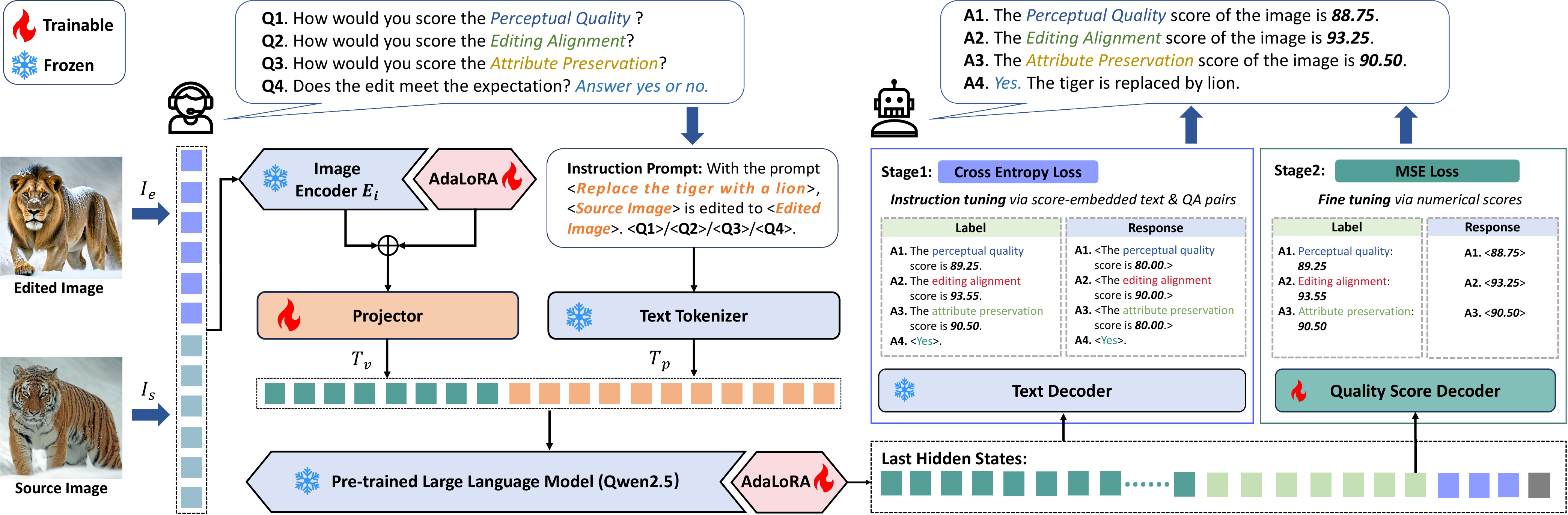}
  \caption{An overview of LMM4Edit. The proposed model takes edited image, source image, and instruction prompt as input, extracts visual and text features, fuses features using a pre-trained LMM, and final outputs scores or answers. We adopt a two-stage training process, of which the first stage is restricted by the text cross-entropy loss, and the second stage is regressed by the score mean-squared-error loss. To enable multi-dimensional evaluation, AdaLoRA \cite{adalora} is applied to both the vision encoder and the LLM for task adaptation.}
  \label{model}
\end{figure*}
\subsection{Subjective Data Processing}
We follow the guidelines outlined in \cite{subject} to identify and exclude outliers, as well as to reject subjects who provide unreliable ratings. An individual rating for an image is considered an outlier if it falls outside $2$ standard deviations (if normal) or $\sqrt{20}$
standard deviations (if not normal) from the mean rating of that image \cite{subject}. A subject is excluded if over $5\%$ of their ratings are outliers. 
As a result, no subject was excluded based on this criterion and $2.26\%$ of the total subjective ratings are removed. 
The remaining valid ratings are converted into Z-scores \cite{subject}, then linearly scaled to the range [0,100]. The final MOS is calculated as follows
\begin{equation}
    z_{ij} = \frac{r_{ij} - \mu_i}{\sigma_i}, \ z_j = \frac{1}{N_j} \sum_{i=1}^{N_j} z_{ij}, \
    MOS_j = \frac{100(z_j + 3)}{6}
\end{equation}
where where $r_{ij}$ is the raw rating given by the i-th subject to the j-th image, $\mu_i$ is the mean rating and $\sigma_i$ is the standard deviation provided by the i-th subject and $N_j$ is the number of valid ratings for the j-th image.
\subsection{Subjective Data Analysis}

Figure~\ref{dataanalysis}(a) shows the MOS distributions, offering an overview of the overall performance of all models across the three dimensions, including perceptual quality, editing alignment, and attribute preservation. Figure~\ref{dataanalysis}(b) displays the distribution of task counts and corresponding averaged MOSs across tasks, emphasizing the diversity of challenges and variations in model performance for different tasks.

We compare various TIE generation models based on MOS ratings of perceptual quality, editing alignment and attribute preservation, and task-specific accuracy performance, as shown in Figure~\ref{allcomparison}. Notably, RFSE \cite{RFSE} achieves strong editing alignment but struggles with attribute preservation, while CDS \cite{CDS} and ZONE \cite{zone} exhibit the opposite trend. Meanwhile, the trend in perceptual quality differs from both editing alignment and attribute preservation, highlighting the need to evaluate TIE models from multiple perspectives. A more detailed analysis of the performance of TIE models across different tasks is presented in Figure~\ref{MOScomparison}.

\begin{figure*}
  \includegraphics[width=0.95\textwidth,height=0.185\textwidth]{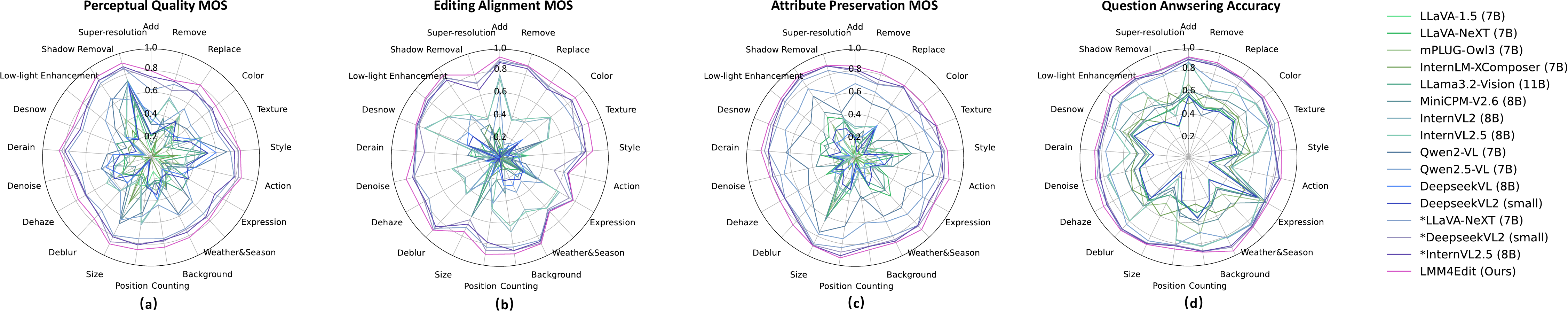}
  \caption{Comparison of quality score correlation with MOSs and QA accuracy of different LMM models in terms of perceptual quality, editing alignment, attribute preservation, and task-specific question answering, respectively.}
  \label{LMM}
\end{figure*}
\begin{table}
\caption{Performance comparisons of quality evaluation methods on EBench-18K from perspectives of perceptual \underline{quality}, editing \underline{alignment}, and attribute \underline{preservation}. 
SRCC ($\rho_s$), KRCC ($\rho_k$), and PLCC ($\rho_p$) metrics are reported. 
$\spadesuit$ Traditional FR IQA metrics, $\heartsuit$ traditional NR IQA metrics, $\clubsuit$ deep learning-based FR IQA methods, $\diamondsuit$ deep learning-based NR IQA methods, \ding{72} vision-language methods, \ding{73} LMM-based models. The fine-tuned results are marked with \raisebox{0.5ex}{\scriptsize \ding{91}}. The best results are highlighted in \mredbf{red}, and the second-best results are highlighted in \mbluebf{blue}.}
\vspace{10pt}
\begin{center}
\centering
\renewcommand\arraystretch{0.95}
\centering
\belowrulesep=0pt
\aboverulesep=0pt
\resizebox{0.48\textwidth}{!}{
\begin{tabular}{l||ccc|ccc|ccc}
\toprule
\noalign{\vspace{1.5pt}}
Dimensions& \multicolumn{3}{c}{Quality} & \multicolumn{3}{c}{Alignment} & \multicolumn{3}{c}{Preservation} \\
\cmidrule(lr){2-4}
\cmidrule(lr){5-7}
\cmidrule(lr){8-10}
\noalign{\vspace{-1.5pt}}
Methods/Metrics & $\rho_s$ & $\rho_k$ & $\rho_p$ &  $\rho_s$ & $\rho_k$ & $\rho_p$ & $\rho_s$ & $\rho_k$ & $\rho_p$  \\
\hline
\noalign{\vspace{1pt}}

$\spadesuit$MSE &0.0257&0.0176&0.2284&0.2178&0.1453&0.0064 &0.4667&0.3251&0.4944 \\
$\spadesuit$PSNR &0.0257&0.0176&0.2134&0.2178&0.1453&0.2742&0.4667&0.3251&0.4999  \\
$\spadesuit$SSIM\cite{SSIM} & 0.0035&0.0007&0.2133&0.1705&0.1132&0.2301&0.4635&0.3217&0.4865 \\
$\spadesuit$FSIM\cite{FSIM} & 0.0469 & 0.0315&0.2487 & 0.2188 & 0.1477 & 0.2883 & 0.6050 & 0.4289 & 0.6174 \\
$\spadesuit$SCSSIM\cite{SCSSIM} & 0.0640&0.0433&0.2646&0.2030&0.1350&0.2670&0.5868&0.4149&0.5938 \\
$\spadesuit$GMSD\cite{GMSD} &0.0099&0.0063&0.0959&0.2208&0.1485&0.2693&0.5272&0.3689&0.5328  \\

\hline
\noalign{\vspace{1pt}}
$\heartsuit$BIQI\cite{BIQI} &0.3002&0.2022&0.3492 &0.1129&0.0755&0.1508 &0.1547&0.1061&0.2346\\
$\heartsuit$DIIVINE\cite{DIIVINE} &0.1429&0.0929&0.3555&0.0471&0.0305&0.1196 &0.0121&0.0083&0.2051  \\
$\heartsuit$BRISQUE\cite{BRISQUE} & 0.3423&0.2360&0.3955&0.1366&0.0923&0.1562&0.1302&0.0883&0.2095\\
$\heartsuit$BLIINDS-II\cite{BLII} & 0.2364&0.1608&0.2878&0.1143&0.0764&0.1243&0.0998&0.0689&0.1476\\
$\heartsuit$NIQE\cite{NIQE} &0.2979&0.2069&0.2453&0.1121&0.0763&0.1080&0.1748&0.1200&0.1926  \\
\hline
\noalign{\vspace{1pt}}
$\clubsuit$LPIPS (alex) \cite{LPIPS} & 0.1832&0.1234&0.2782&0.2222&0.1489&0.2992&0.7395&0.5478&0.7594 \\
$\clubsuit$LPIPS (vgg) \cite{LPIPS} & 0.1643&0.1101&0.1902&0.2166&0.1452&0.2632&0.7248&0.5326&0.7430 \\
$\clubsuit$ST-LPIPS (alex) \cite{STLPIPS}& 0.0052&0.0045&0.0521&0.2123&0.1432&0.1325&0.4996&0.3463&0.3943  \\
$\clubsuit$ST-LPIPS (vgg) \cite{STLPIPS} &  0.1048&0.0696&0.0510&0.2403&0.1619&0.1830&0.4161&0.2841&0.3589\\
$\clubsuit$CVRKD\raisebox{0.5ex}{\scriptsize \ding{91}} \cite{CVRKD} &  0.7935&0.5991&0.8106&0.4661&0.3170&0.4806&0.7864&0.5917&0.8081\\
$\clubsuit$AHIQ\raisebox{0.5ex}{\scriptsize \ding{91}}  \cite{AHIQ} & 0.8183&0.6241&0.8324&0.5249&0.3679&0.5452& 0.8365&0.6457&0.8515 \\

\hline
\noalign{\vspace{1pt}}
$\diamondsuit$CNNIQA\raisebox{0.5ex}{\scriptsize \ding{91}}  \cite{CNN} & 0.6336 & 0.4491 & 0.6552 & 0.2166 & 0.1471 & 0.2229 & 0.3075 & 0.2869 & 0.2075 \\
$\diamondsuit$WaDIQaM\raisebox{0.5ex}{\scriptsize \ding{91}}  \cite{Wa} & 0.6501&0.4647&0.6797&0.2412&0.1640&0.2626&0.3590&0.2449&0.3926 \\
$\diamondsuit$NIMA\raisebox{0.5ex}{\scriptsize \ding{91}} 
 \cite{NIMA} & 0.5748&0.4023&0.5928&0.2070&0.1394&0.2272&0.3368&0.2287&0.2907 \\
$\diamondsuit$DBCNN\raisebox{0.5ex}{\scriptsize \ding{91}}  \cite{DBCNN}& 0.7646&0.5690&0.7805&0.3248&0.2264&0.3731&0.6241&0.4500&0.6782 \\
$\diamondsuit$HyperIQA\raisebox{0.5ex}{\scriptsize \ding{91}}  \cite{Hyper} &0.6543&0.4692&0.6683& 0.2283&0.1566&0.2305&0.2810&0.1891&0.2963 \\
$\diamondsuit$MANIQA\raisebox{0.5ex}{\scriptsize \ding{91}}  \cite{MANIQA} &0.8050 &0.6136 &0.8171 &0.3432&0.2661&0.3765 &0.6529&0.4716&0.7041\\
$\diamondsuit$CLIPIQA\raisebox{0.5ex}{\scriptsize \ding{91}}  \cite{CLIPIQA}& 0.7721 &0.5700&0.7711&0.3187&0.2899&0.3334&0.5640&0.3957&0.5938 \\
$\diamondsuit$TOPIQ\raisebox{0.5ex}{\scriptsize \ding{91}}  \cite{TOPIQ}& 0.7936&0.6021&0.8054&0.3641&0.2536&0.3848 &0.6320&0.6692&0.4565\\
$\diamondsuit$Q-Align\raisebox{0.5ex}{\scriptsize \ding{91}}  \cite{qalign}&0.8180&0.6285&0.8014&0.4961&0.3684&0.4994& 0.7046&0.5188&0.7321\\
\hline
\noalign{\vspace{1pt}}

\ding{72}CLIPScore \cite{clipscore} & 0.2181 & 0.1467 & 0.2243 & 0.2152 & 0.1449 & 0.2209 & 0.2325 & 0.1586 & 0.2581 \\
\ding{72}BLIPScore \cite{blip}& 0.2721 & 0.1876 & 0.2880 & 0.1852 & 0.1240 & 0.1721 & 0.3625 & 0.2475 & 0.3870 \\
\ding{72}ImageReward \cite{imagereward} & 0.3991 & 0.2764 & 0.4351 & 0.2875 & 0.1978 & 0.3198 & 0.4033 & 0.2779 & 0.4662 \\
\ding{72}PickScore \cite{pickscore}& 0.2483 & 0.1666 & 0.2889 & 0.3627 & 0.2482 & 0.3297 & 0.1357 & 0.0874 & 0.2046 \\
\ding{72}HPSv2 \cite{HPS} & 0.5376 & 0.3702 & 0.5584 & 0.3189 & 0.2197 & 0.3500 & 0.3764 & 0.2543 & 0.4241 \\
\ding{72}LLaVAScore \cite{llavascore}& 0.2981 & 0.2070 & 0.3670 & 0.2590 & 0.1726 & 0.2786 & 0.3047 & 0.2088 & 0.3978 \\
\ding{72}VQAScore \cite{vqa}& 0.3014 & 0.2050 & 0.3162 & 0.2839 & 0.1898 & 0.2695& 0.2185 & 0.1444 & 0.2537  \\

\hline
\noalign{\vspace{1pt}}
\ding{73}LLaVA-1.5 (7B) \cite{llava} & 0.3088 & 0.2070 & 0.3670 & 0.1185 & 0.0969 & 0.1576 & 0.0447 & 0.0396 & 0.0494 \\
\ding{73}mPLUG-Owl3 (7B) \cite{mplug}& 0.3116 & 0.2244 & 0.0265 & 0.0597 & 0.0422 & 0.0174 & 0.1547 & 0.1091 & 0.0747 \\
\ding{73}InternLM-XComposer (7B) \cite{internlmxcomposer}& 0.0637 & 0.0471 & 0.1677 & 0.0369 & 0.0276 & 0.0502 & 0.1047 & 0.0795 & 0.1351 \\
\ding{73}LLama3.2-Vision (11B) \cite{llama} & 0.5383 & 0.4187 & 0.4317 & 0.1272 & 0.1039 & 0.0922 & 0.3832 & 0.2980 & 0.3812 \\
\ding{73}MiniCPM-V2.6 (8B) \cite{minicpm}& 0.5343 & 0.4321 & 0.5101 & 0.1781 & 0.1427 & 0.1720 & 0.4224 & 0.3308 & 0.3470 \\
\ding{73}DeepSeekVL (7B) \cite{deepseekvl} & 0.6786 & 0.4866 & 0.7041  & 0.1727 & 0.1350 & 0.1893 & 0.2367 & 0.1853 & 0.2122 \\
\ding{73}InternVL2 (8B) \cite{internvl}& 0.6329 & 0.4432 & 0.6664 & 0.4141 & 0.3137 & 0.4057 & 0.2304 & 0.1773 & 0.1684 \\
\ding{73}Qwen2-VL (7B) \cite{qwenvl} & 0.6786 & 0.4866 & 0.7041 & 0.1937 & 0.1427 & 0.1720 & 0.5478 & 0.3981 & 0.5200\\ 
\ding{73}LLaVA-NeXT (8B)\raisebox{0.5ex}{\scriptsize \ding{91}} \cite{llava-n}& 0.8601 & 0.7251 & 0.8657 &0.8125 & 0.6714 & 0.8241 & 0.8517 & 0.7206 & 0.8494 \\
\ding{73}DeepSeekVL2 (small)\raisebox{0.5ex}{\scriptsize \ding{91}} \cite{deepseekvl2} & 0.8674 & \mbluebf{0.7303} & 0.8665 & \mbluebf{0.8280} & \mbluebf{0.6893} & \mbluebf{0.8390} & 0.8778 & 0.7500 & 0.8819 \\
\ding{73}InternVL2.5 (8B)\raisebox{0.5ex}{\scriptsize \ding{91}} \cite{internvl2}& \mbluebf{0.8836} & 0.7223 & \mbluebf{0.8861} & 0.8207 & 0.6809 & 0.8387 & \mbluebf{0.8841} & \mbluebf{0.7571} & \mbluebf{0.8949} \\
\hline
\noalign{\vspace{1pt}}
\rowcolor{gray!20}  
\textbf{LMM4Edit (Ours)} & \mredbf{0.9136} & \mredbf{0.7432} & \mredbf{0.9189} & \mredbf{0.8830} & \mredbf{0.7080} & \mredbf{0.8898} & \mredbf{0.9048} & \mredbf{0.7837} & \mredbf{0.9176} \\
\rowcolor{gray!20}  
Improvement & +3.28\%& +1.77\%& +3.57\%&+6.64\%&+2.71\%&+6.05\%&+2.34\%&+3.51\%&+2.54\%\\
\bottomrule

\end{tabular}}
\vspace{-8pt}
\label{performances}
\end{center}
\end{table}
\begin{table*}[tb]
\renewcommand{\arraystretch}{0.85}
\centering
\caption{Performance comparison of LMMs across different editing tasks in EBench-18K. High-level tasks include: add, remove, replace, color, texture, style, action, expression, weather$\&$season, background, counting, position, and size. Low-level tasks include: deblur, dehaze, denoise, derain, desnow, low-light enhancement, shadow removal, and super-resolution. We report the SRCC between predicted scores from evaluation models and MOSs for perceptual quality ($\rho_p$), editing alignment ($\rho_e$), and attribute preservation ($\rho_a$), along with QA accuracy ($Acc$). \raisebox{0.5ex}{\scriptsize \ding{91}} denotes fine-tuned models. The best results are highlighted in \mredbf{red}, while the second-best results are highlighted in \mbluebf{blue}.
}
\label{comparison_task}
\setlength{\belowcaptionskip}{-0.01cm}
\belowrulesep=0pt
\aboverulesep=0pt
 \resizebox{1\textwidth}{!}{
\begin{tabular}{l|| ccc:c| ccc:c| ccc:c| ccc:c| ccc:c| ccc:c}
\toprule
\noalign{\vspace{0.5pt}}
Editing Tasks & \multicolumn{4}{c}{Add} & \multicolumn{4}{c}{Remove} & \multicolumn{4}{c}{Replace}& \multicolumn{4}{c}{Color}& \multicolumn{4}{c}{Texture}& \multicolumn{4}{c}{Style} \\
\cmidrule(lr){2-5}
\cmidrule(lr){6-9}
\cmidrule(lr){10-13}
\cmidrule(lr){14-17}
\cmidrule(lr){18-21}
\cmidrule(lr){22-25}
Methods/Metrics 
& $\rho_p$ & $\rho_e$ & $\rho_a$ & $Acc$
&$\rho_p$ & $\rho_e$ & $\rho_a$&$Acc$
& $\rho_p$ & $\rho_e$ & $\rho_a$&$Acc$
& $\rho_p$ & $\rho_e$ & $\rho_a$&$Acc$
& $\rho_p$ & $\rho_e$ & $\rho_a$& $Acc$
&$\rho_p$ & $\rho_e$ & $\rho_a$&$Acc$ \\
\hline
\noalign{\vspace{1pt}}
LLaVA-1.5 (7B) \cite{llava} &0.061&0.269 &0.098 &0.559&0.163&0.099 &0.023 &0.463&0.056&0.024 &0.072 &0.507&0.073&0.088 &0.050 &0.529&0.044&0.125 &0.097 &0.434&0.277&0.264 &0.028&0.449\\

LLaVA-NeXT (8B) \cite{llava-n} &0.196&0.106 &0.163 &0.566&0.266&0.030 &0.139 &0.478&0.348&0.130 &0.335 &0.522&0.235&0.029 &0.174 &0.544&0.221&0.023 &0.148 &0.412&0.491&0.023 &0.502&0.493\\

mPLUG-Owl3 (7B) \cite{mplug} &0.148&0.010 &0.206 &0.676&0.139&0.020 &0.209 &0.485&0.463&0.014 &0.212 &0.544&0.315&0.014 &0.140 &0.596&0.151&0.056 &0.216 &0.449&0.180&0.068 &0.089&0.574\\

InternLM-XComposer (7B) \cite{internlmxcomposer} &0.043&0.132 &0.081 &0.493&0.189&0.077 &0.125 &0.596&0.018&0.051 &0.051 &0.588&0.184&0.057 &0.047 &0.581&0.024&0.014 &0.070 &0.537&0.247&0.070&0.020 &0.574\\

LLama3.2-Vision (11B) \cite{llama} &0.249&0.282 &0.217 &0.596&0.249&0.101 &0.131 &0.478&0.333&0.039 &0.338 &0.537&0.155&0.075 &0.077 &0.522&0.221&0.104 &0.157 &0.397&0.491&0.246 &0.349&0.500\\

MiniCPM-V2.6 (8B) \cite{minicpm} &0.113&0.080 &0.187 &0.625&0.379&0.093 &0.256 &0.574&0.231&0.053 &0.097 &0.632&0.308&0.187 &0.251 &0.676&0.271&0.039 &0.438 &0.588&0.526&0.018 &0.520&0.485\\

InternVL2 (8B) \cite{internvl} &0.316&0.764 &0.287 &0.882&0.576&0.372 &0.305 &0.640&0.470&0.396 &0.323 &0.721&0.135&0.587 &0.094 &0.797&0.494&0.468 &0.247 &0.801&0.328&0.121 &0.355&0.647\\

InternVL2.5 (8B) \cite{internvl2} &0.321&0.726 &0.268 &0.875&0.556&0.360 &0.300 &0.647&0.473&0.436 &0.348 &0.721&0.198&0.599 &0.092 &0.790&0.509&0.479 &0.249 &0.794&0.326&0.151 &0.283&0.632\\

Qwen2-VL (7B) \cite{qwenvl} &0.306&0.030 &0.603 &0.581&0.317&0.013 &0.520 &0404&0.388&0.142 &0.534 &0.551&0.422&0.067 &0.474 &0.559&0446 &0.418 &0.423 &0.471&0.700&0.058 &0.649&0.463\\

Qwen2.5-VL (7B) \cite{qwenvl2} &0.420&0.612 &0.683 &0.875&0.325&0.127 &0.632 &0.743&0.542&0.164 &0.588 &0.779&0.469&0.178 &0.636 &0.699&0486 &0.157 &0.682 &0.779&0.457&0.191 &0.726&0.772\\

DeepSeekVL (7B) \cite{deepseekvl} &0.354&0.076 &0.187 &0.559&0.302&0.078 &0.201 &0.463&0.289&0.113 &0.360 &0.779&0.297&0.272 &0.134 &0.529&0.449&0.057 &0.121 &0.434&0.602&0.164 &0.204&0.449\\

DeepSeekVL2 (small) \cite{deepseekvl2} &0.213&0.053 &0.185 &0.559&0.265&0.052 &0.056 &0.463&0.406&0.001 &0.314 &0.507&0.256&0.213 &0.137 &0.529&0.389&0.087 &0.083 &0.434&0.525&0.132 &0.350&0.449\\

\hline
\raisebox{0.5ex}{\scriptsize \ding{91}}LLaVA-NeXT (8B) \cite{llava-n} &0.621&0.886&0.761&\mbluebf{0.919}&0.704&\mbluebf{0.881} &0.706 &0.897&0.726&0.786 &0.783 &0.875&0.686&0.761 &\mbluebf{0.741} &0.868&0.645&0.777 &0.685 &\mbluebf{0.853}&0.733&0.760 &0.817&0.794\\

\raisebox{0.5ex}{\scriptsize \ding{91}}DeepSeekVL2 (small) \cite{deepseekvl2} &0.727&\mbluebf{0.905}&0.798&0.897&0.645&0.880 &0.712 &\mbluebf{0.897}&\mbluebf{0.749}&\mbluebf{0.799} &0.785 &0.860&0.707&0.844 &0.735 &\mbluebf{0.875}&\mbluebf{0.707}&0.739 &0.762 &0.824&\mbluebf{0.807}&0.791 &\mbluebf{0.828}&\mbluebf{0.846}\\

\raisebox{0.5ex}{\scriptsize \ding{91}}InternVL2.5 (8B) \cite{internvl2} &\mbluebf{0.743}&0.877&\mbluebf{0.816}&0.904&\mbluebf{0.731}&0.859 &\mbluebf{0.750} &0.882&0.714&0.767 &\mbluebf{0.790} &\mbluebf{0.881}&\mbluebf{0.722}&\mbluebf{0.850} &0.708 &0.868&0.674&\mbluebf{0.813} &\mbluebf{0.800} &0.809&0.771&\mbluebf{0.793} &0.795&0.801\\

\rowcolor{gray!20}  
\raisebox{0.5ex}{\scriptsize \ding{91}}LMM4Edit (Ours) 
& \mredbf{0.797} & \mredbf{0.928} & \mredbf{0.844} & \mredbf{0.926} 
& \mredbf{0.739} & \mredbf{0.885} & \mredbf{0.818} & \mredbf{0.912} 
& \mredbf{0.808} & \mredbf{0.843} & \mredbf{0.804} & \mredbf{0.882} 
& \mredbf{0.762} & \mredbf{0.889} & \mredbf{0.801} & \mredbf{0.890} 
& \mredbf{0.758} & \mredbf{0.876} & \mredbf{0.807} & \mredbf{0.890} 
& \mredbf{0.826} & \mredbf{0.859} & \mredbf{0.853} & \mredbf{0.860} \\

 \noalign{\vspace{-1.5pt}}
\bottomrule
\end{tabular}}

 \resizebox{1\textwidth}{!}{
\begin{tabular}{l|| ccc:c| ccc:c| ccc:c| ccc:c| ccc:c| ccc:c}
\noalign{\vspace{3pt}}
\toprule
\noalign{\vspace{0.5pt}}
Editing Tasks& \multicolumn{4}{c}{Action} & \multicolumn{4}{c}{Expression} & \multicolumn{4}{c}{Weather$\&$Season}& \multicolumn{4}{c}{Background}& \multicolumn{4}{c}{Counting}& \multicolumn{4}{c}{Position} \\
\cmidrule(lr){2-5}
\cmidrule(lr){6-9}
\cmidrule(lr){10-13}
\cmidrule(lr){14-17}
\cmidrule(lr){18-21}
\cmidrule(lr){22-25}
Methods/Metrics 
& $\rho_p$ & $\rho_e$ & $\rho_a$ & $Acc$
&$\rho_p$ & $\rho_e$ & $\rho_a$&$Acc$
& $\rho_p$ & $\rho_e$ & $\rho_a$&$Acc$
& $\rho_p$ & $\rho_e$ & $\rho_a$&$Acc$
& $\rho_p$ & $\rho_e$ & $\rho_a$& $Acc$
&$\rho_p$ & $\rho_e$ & $\rho_a$&$Acc$ \\
\hline
\noalign{\vspace{1pt}}
LLaVA-1.5 (7B) \cite{llava} & 0.042 & 0.075 & 0.075 & 0.191 & 0.103 & 0.387 & 0.150 & 0.728 & 0.274 & 0.044 & 0.028 & 0.272 & 0.271 & 0.051 & 0.118 & 0.441 & 0.131 & 0.126 & 0.021 & 0.559 & 0.266 & 0.014 & 0.068 & 0.404 \\
LLaVA-NeXT (8B) \cite{llava-n} & 0.127 & 0.019 & 0.159 & 0.228 & 0.389 & 0.215 & 0.347 & 0.743 & 0.415 & 0.149 & 0.481 & 0.375 & 0.237 & 0.101 & 0.311 & 0.485 & 0.247 & 0.056 & 0.088 & 0.515 & 0.250 & 0.094 & 0.28 & 0.434 \\
mPLUG-Owl3 (7B) \cite{mplug} & 0.266 & 0.116 & 0.180 & 0.228 & 0.076 & 0.086 & 0.082 & 0.809 & 0.233 & 0.005 & 0.203 & 0.463 & 0.267 & 0.032 & 0.147 & 0.456 & 0.307 & 0.012 & 0.126 & 0.713 & 0.256 & 0.013 & 0.169 & 0.603 \\
InternLM-XComposer (7B) \cite{internlmxcomposer} &0.077 & 0.081 & 0.005 & 0.382 & 0.213 & 0.181 & 0.089 & 0.596 & 0.205 & 0.013 & 0.002 & 0.610 & 0.303 & 0.124 & 0.072 & 0.566 & 0.004 & 0.102 & 0.008 & 0.434 & 0.151 & 0.130 & 0.023 & 0.485 \\
LLama3.2-Vision (11B) \cite{llama} &0.128 & 0.046 & 0.097 & 0.228 & 0.389 & 0.387 & 0.304 & 0.743 & 0.415 & 0.044 & 0.397 & 0.375 & 0.237 & 0.051 & 0.243 & 0.485 & 0.215 & 0.126 & 0.062 & 0.515 & 0.282 & 0.014 & 0.205 & 0.419 \\
MiniCPM-V2.6 (8B) \cite{minicpm} & 0.311 & 0.143 & 0.358 & 0.404 & 0.106 & 0.224 & 0.171 & 0.831 & 0.194 & 0.242 & 0.410 & 0.574 & 0.333 & 0.105 & 0.360 & 0.632 & 0.218 & 0.332 & 0.381 & 0.625 & 0.298 & 0.203 & 0.209 & 0.515 \\
InternVL2 (8B) \cite{internvl} & 0.342 & 0.048 & 0.339 & 0.404 & 0.215 & 0.635 & 0.159 & 0.760 & 0.132 & 0.634 & 0.136 & 0.838 & 0.281 & 0.528 & 0.104 & 0.816 & 0.403 & 0.691 & 0.153 & 0.824 & 0.631 & 0.488 & 0.340 & 0.765 \\
InternVL2.5 (8B) \cite{internvl2} & 0.316 & 0.089 & 0.305 & 0.412 & 0.187 & 0.618 & 0.180 & 0.760 & 0.224 & 0.632 & 0.111 & 0.846 & 0.271 & 0.552 & 0.109 & 0.816 & 0.401 & 0.637 & 0.174 & 0.831 & 0.61 & 0.471 & 0.341 & 0.772 \\
Qwen2-VL (7B) \cite{qwenvl} & 0.420 & 0.244 & 0.744 & 0.338 & 0.169 & 0.393 & 0.542 & 0.750 & 0.520 & 0.022 & 0.628 & 0.397 & 0.577 & 0.035 & 0.598 & 0.493 & 0.234 & 0.108 & 0.357 & 0.588 & 0.510 & 0.010 & 0.637 & 0.544 \\
Qwen2.5-VL (7B) \cite{qwenvl2} & 0.370 & 0.178 & 0.636 & 0.743 & 0.347 & 0.462 & 0.720 & 0.809 & 0.561 & 0.292 & 0.708 & 0.846 & 0.591 & 0.352 & 0.654 & 0.824 & 0.442 & 0.281 & 0.607 & 0.699 & 0.547 & 0.425 & 0.754 & 0.816 \\
DeepSeekVL (7B) \cite{deepseekvl} & 0.415 & 0.201 & 0.269 & 0.191 & 0.315 & 0.268 & 0.105 & 0.728 & 0.151 & 0.210 & 0.118 & 0.279 & 0.232 & 0.281 & 0.221 & 0.441 & 0.393 & 0.262 & 0.093 & 0.566 & 0.233 & 0.090 & 0.27 & 0.412 \\
DeepSeekVL2 (small) \cite{deepseekvl2} & 0.338 & 0.111 & 0.279 & 0.191 & 0.397 & 0.243 & 0.158 & 0.728 & 0.206 & 0.273 & 0.224 & 0.279 & 0.212 & 0.222 & 0.288 & 0.441 & 0.350 & 0.213 & 0.051 & 0.566 & 0.272 & 0.019 & 0.296 & 0.412 \\
\hline
\noalign{\vspace{1pt}}
\raisebox{0.5ex}{\scriptsize \ding{91}}LLaVA-NeXT (8B) \cite{llava-n} & 0.784 & 0.603 & 0.782 & 0.853 & 0.707 & 0.776 & 0.740 & 0.809 & 0.708 & 0.658 & 0.849 & 0.868 & 0.727 & 0.845 & 0.795 & 0.919 & 0.757 & 0.831 & 0.788 & 0.875 & 0.758 & 0.833 & 0.836 & 0.816 \\
\raisebox{0.5ex}{\scriptsize \ding{91}}DeepSeekVL2 (small) \cite{deepseekvl2} & \mbluebf{0.828} & 0.514 & \mbluebf{0.840} & 0.816 & \mbluebf{0.753} & 0.739 & \mbluebf{0.835} & \mbluebf{0.831} & \mbluebf{0.754} & 0.648 & 0.858 & \mbluebf{0.868} & 0.738 & \mbluebf{0.872} & 0.836 & \mbluebf{0.919} & \mbluebf{0.781} & \mbluebf{0.864} & 0.825 & 0.868 & 0.802 & \mbluebf{0.864} & 0.878 & 0.816 \\
\raisebox{0.5ex}{\scriptsize \ding{91}}InternVL2.5 (8B) \cite{internvl2} & 0.793 & \mbluebf{0.640} & 0.834 & \mbluebf{0.860} & 0.676 & \mbluebf{0.786} & 0.789 & 0.809 & 0.744 & \mbluebf{0.670} & \mbluebf{0.849} & 0.860 & \mbluebf{0.765} & 0.856 & \mbluebf{0.851} & 0.904 & 0.770 & 0.863 & \mbluebf{0.827} & \mbluebf{0.875} & \mbluebf{0.815} & 0.786 & \mbluebf{0.912} & \mbluebf{0.824} \\
\rowcolor{gray!20}  
\raisebox{0.5ex}{\scriptsize \ding{91}}LMM4Edit (Ours) 
& \mredbf{0.852} & \mredbf{0.648} & \mredbf{0.887} & \mredbf{0.875} 
& \mredbf{0.762} & \mredbf{0.781} & \mredbf{0.838} & \mredbf{0.846} 
& \mredbf{0.782} & \mredbf{0.674} & \mredbf{0.901} & \mredbf{0.875} 
& \mredbf{0.818} & \mredbf{0.878} & \mredbf{0.865} & \mredbf{0.956} 
& \mredbf{0.836} & \mredbf{0.895} & \mredbf{0.857} & \mredbf{0.882} 
& \mredbf{0.859} & \mredbf{0.899} & \mredbf{0.932} & \mredbf{0.853} \\

 \noalign{\vspace{-1.5pt}}
\bottomrule
\end{tabular}}

 \resizebox{1\textwidth}{!}{
\begin{tabular}{l|| ccc:c| ccc:c| ccc:c| ccc:c| ccc:c| ccc:c}
\noalign{\vspace{3pt}}
\toprule
\noalign{\vspace{0.5pt}}
Editing Tasks & \multicolumn{4}{c}{Size} & \multicolumn{4}{c}{Deblur} & \multicolumn{4}{c}{Dehaze}& \multicolumn{4}{c}{Denoise}& \multicolumn{4}{c}{Derain}& \multicolumn{4}{c}{Desnow} \\
\cmidrule(lr){2-5}
\cmidrule(lr){6-9}
\cmidrule(lr){10-13}
\cmidrule(lr){14-17}
\cmidrule(lr){18-21}
\cmidrule(lr){22-25}
Methods/Metrics 
& $\rho_p$ & $\rho_e$ & $\rho_a$ & $Acc$
&$\rho_p$ & $\rho_e$ & $\rho_a$&$Acc$
& $\rho_p$ & $\rho_e$ & $\rho_a$&$Acc$
& $\rho_p$ & $\rho_e$ & $\rho_a$&$Acc$
& $\rho_p$ & $\rho_e$ & $\rho_a$& $Acc$
&$\rho_p$ & $\rho_e$ & $\rho_a$&$Acc$ \\
\hline
\noalign{\vspace{1pt}}
LLaVA-1.5 (7B) \cite{llava} & 0.279 & 0.500 & 0.072 & 0.140 & 0.106 & 0.153 & 0.069 & 0.529 & 0.030 & 0.108 & 0.061 & 0.475 & 0.074 & 0.051 & 0.048 & 0.466 & 0.091 & 0.271 & 0.045 & 0.319 & 0.059 & 0.262 & 0.144 & 0.534 \\
LLaVA-NeXT (8B) \cite{llava-n} & 0.167 & 0.080 & 0.602 & 0.382 & 0.235 & 0.048 & 0.155 & 0.538 & 0.303 & 0.047 & 0.306 & 0.487 & 0.251 & 0.078 & 0.114 & 0.466 & 0.247 & 0.097 & 0.316 & 0.361 & 0.342 & 0.142 & 0.346 & 0.563 \\
mPLUG-Owl3 (7B) \cite{mplug} & 0.454 & 0.342 & 0.537 & 0.471 & 0.131 & 0.014 & 0.177 & 0.605 & 0.354 & 0.089 & 0.138 & 0.517 & 0.175 & 0.023 & 0.188 & 0.525 & 0.137 & 0.049 & 0.107 & 0.382 & 0.072 & 0.100 & 0.081 & 0.647 \\
InternLM-XComposer (7B) \cite{internlmxcomposer} & 0.474 & 0.196 & 0.122 & 0.596 & 0.074 & 0.041 & 0.096 & 0.525 & 0.068 & 0.015 & 0.065 & 0.597 & 0.013 & 0.072 & 0.051 & 0.542 & 0.191 & 0.078 & 0.022 & 0.479 & 0.227 & 0.156 & 0.063 & 0.597 \\
LLama3.2-Vision (11B) \cite{llama} & 0.194 & 0.490 & 0.453 & 0.382 & 0.217 & 0.139 & 0.173 & 0.550 & 0.301 & 0.057 & 0.263 & 0.492 & 0.222 & 0.053 & 0.070 & 0.458 & 0.247 & 0.224 & 0.241 & 0.361 & 0.342 & 0.258 & 0.305 & 0.563 \\
MiniCPM-V2.6 (8B) \cite{minicpm} & 0.637 & 0.600 & 0.646 & 0.331 & 0.251 & 0.093 & 0.190 & 0.613 & 0.164 & 0.058 & 0.132 & 0.601 & 0.347 & 0.025 & 0.370 & 0.601 & 0.352 & 0.143 & 0.324 & 0.462 & 0.080 & 0.283 & 0.258 & 0.681 \\
InternVL2 (8B) \cite{internvl} & 0.198 & 0.188 & 0.310 & 0.213 & 0.429 & 0.460 & 0.288 & 0.798 & 0.444 & 0.365 & 0.193 & 0.744 & 0.322 & 0.441 & 0.030 & 0.773 & 0.207 & 0.112 & 0.270 & 0.538 & 0.044 & 0.748 & 0.029 & 0.753 \\
InternVL2.5 (8B) \cite{internvl2} & 0.240 & 0.005 & 0.323 & 0.213 & 0.420 & 0.446 & 0.271 & 0.798 & 0.473 & 0.382 & 0.203 & 0.744 & 0.333 & 0.454 & 0.044 & 0.761 & 0.190 & 0.133 & 0.224 & 0.538 & 0.096 & 0.733 & 0.047 & 0.757 \\
Qwen2-VL (7B) \cite{qwenvl} & 0.679 & 0.201 & 0.760 & 0.301 & 0.312 & 0.003 & 0.572 & 0.500 & 0.382 & 0.006 & 0.494 & 0.513 & 0.471 & 0.121 & 0.444 & 0.513 & 0.406 & 0.296 & 0.639 & 0.399 & 0.240 & 0.263 & 0.547 & 0.576 \\
Qwen2.5-VL (7B) \cite{qwenvl2} & 0.637 & 0.234 & 0.779 & 0.853 & 0.418 & 0.404 & 0.670 & 0.840 & 0.423 & 0.033 & 0.608 & 0.739 & 0.460 & 0.138 & 0.672 & 0.735 & 0.284 & 0.065 & 0.616 & 0.765 & 0.358 & 0.427 & 0.694 & 0.724 \\
DeepSeekVL (7B) \cite{deepseekvl} & 0.044 & 0.185 & 0.051 & 0.147 & 0.325 & 0.060 & 0.206 & 0.529 & 0.316 & 0.083 & 0.250 & 0.475 & 0.424 & 0.134 & 0.103 & 0.466 & 0.351 & 0.012 & 0.177 & 0.319 & 0.069 & 0.323 & 0.056 & 0.538 \\
DeepSeekVL2 (small) \cite{deepseekvl2} & 0.019 & 0.006 & 0.269 & 0.147 & 0.219 & 0.065 & 0.150 & 0.529 & 0.393 & 0.012 & 0.211 & 0.475 & 0.356 & 0.118 & 0.106 & 0.466 & 0.288 & 0.034 & 0.249 & 0.319 & 0.147 & 0.308 & 0.140 & 0.538 \\
\hline
\noalign{\vspace{1pt}}
\raisebox{0.5ex}{\scriptsize \ding{91}}LLaVA-NeXT (8B) \cite{llava-n} & 0.797 & \mbluebf{0.712} & 0.880 & 0.882 & 0.669 & 0.886 & 0.758 & \mbluebf{0.912} & 0.690 & 0.753 & 0.747 & 0.866 & 0.659 & 0.809 & 0.705 & \mbluebf{0.866} & 0.764 & 0.774 & 0.803 & 0.819 & 0.678 & 0.811 & 0.778 & 0.832 \\
\raisebox{0.5ex}{\scriptsize \ding{91}}DeepSeekVL2 (small) \cite{deepseekvl2} & \mbluebf{0.863} & 0.633 & 0.892 & 0.875 & 0.711 & \mbluebf{0.901} & 0.774 & 0.899 & \mbluebf{0.698} & \mbluebf{0.803} & \mbluebf{0.749} & 0.857 & \mbluebf{0.710} & 0.810 & 0.756 & 0.840 & \mbluebf{0.826} & 0.694 & \mbluebf{0.838} & 0.840 & \mbluebf{0.755} & 0.798 & \mbluebf{0.834} & \mbluebf{0.845} \\
\raisebox{0.5ex}{\scriptsize \ding{91}}InternVL2.5 (8B) \cite{internvl2} & 0.812 & 0.679 & \mbluebf{0.898} & \mbluebf{0.882} & \mbluebf{0.717} & 0.867 & \mbluebf{0.780} & 0.899 & 0.692 & 0.792 & 0.748 & \mbluebf{0.870} & 0.701 & \mbluebf{0.823} & \mbluebf{0.778} & 0.832 & 0.796 & \mbluebf{0.782} & 0.802 & \mbluebf{0.832} & 0.702 & \mbluebf{0.820} & 0.811 & 0.815 \\
\rowcolor{gray!20}  
\raisebox{0.5ex}{\scriptsize \ding{91}}LMM4Edit (Ours) 
& \mredbf{0.884} & \mredbf{0.755} & \mredbf{0.899} & \mredbf{0.890} 
& \mredbf{0.769} & \mredbf{0.909} & \mredbf{0.843} & \mredbf{0.916} 
& \mredbf{0.775} & \mredbf{0.857} & \mredbf{0.798} & \mredbf{0.903} 
& \mredbf{0.767} & \mredbf{0.885} & \mredbf{0.807} & \mredbf{0.878} 
& \mredbf{0.849} & \mredbf{0.790} & \mredbf{0.871} & \mredbf{0.866} 
& \mredbf{0.772} & \mredbf{0.830} & \mredbf{0.856} & \mredbf{0.849} \\
 \noalign{\vspace{-1.5pt}}
\bottomrule

\end{tabular}}

 \resizebox{1\textwidth}{!}{
\begin{tabular}{l|| ccc:c| ccc:c| ccc:c| ccc:c| ccc:c| ccc:c}
\noalign{\vspace{3pt}}
\toprule
\noalign{\vspace{0.5pt}}
Editing Tasks & \multicolumn{4}{c}{Low-light Enhancement} & \multicolumn{4}{c}{Shadow Removal} & \multicolumn{4}{c}{Super-resolution}& \multicolumn{4}{c}{Overall High-level}& \multicolumn{4}{c}{Overall Low-level}& \multicolumn{4}{c}{Overall} \\
\cmidrule(lr){2-5}
\cmidrule(lr){6-9}
\cmidrule(lr){10-13}
\cmidrule(lr){14-17}
\cmidrule(lr){18-21}
\cmidrule(lr){22-25}
Methods/Metrics 
& $\rho_p$ & $\rho_e$ & $\rho_a$ & $Acc$
&$\rho_p$ & $\rho_e$ & $\rho_a$&$Acc$
& $\rho_p$ & $\rho_e$ & $\rho_a$&$Acc$
& $\rho_p$ & $\rho_e$ & $\rho_a$&$Acc$
& $\rho_p$ & $\rho_e$ & $\rho_a$& $Acc$
&$\rho_p$ & $\rho_e$ & $\rho_a$&$Acc$ \\
\hline
\noalign{\vspace{1pt}}
LLaVA-1.5 (7B) \cite{llava} & 0.203 & 0.075 & 0.149 & 0.450 & 0.265 & 0.079 & 0.019 & 0.395 & 0.415 & 0.176 & 0.117 & 0.294 & 0.130 & 0.048 & 0.030 & 0.437 & 0.150 & 0.301 & 0.004 & 0.422 & 0.309 & 0.119 & 0.045 & 0.429 \\
LLaVA-NeXT (8B) \cite{llava-n} & 0.246 & 0.220 & 0.307 & 0.521 & 0.235 & 0.013 & 0.452 & 0.475 & 0.601 & 0.082 & 0.376 & 0.475 & 0.261 & 0.068 & 0.296 & 0.475 & 0.304 & 0.334 & 0.232 & 0.600 & 0.559 & 0.151 & 0.419 & 0.540 \\
mPLUG-Owl3 (7B) \cite{mplug} & 0.260 & 0.024 & 0.178 & 0.559 & 0.429 & 0.226 & 0.368 & 0.668 & 0.237 & 0.027 & 0.050 & 0.508 & 0.224 & 0.043 & 0.193 & 0.544 & 0.342 & 0.086 & 0.113 & 0.548 & 0.312 & 0.060 & 0.155 & 0.546 \\
InternLM-XComposer (7B) \cite{internlmxcomposer} & 0.209 & 0.095 & 0.006 & 0.563 & 0.303 & 0.251 & 0.066 & 0.382 & 0.057 & 0.324 & 0.038 & 0.622 & 0.053 & 0.032 & 0.055 & 0.541 & 0.245 & 0.033 & 0.079 & 0.533 & 0.064 & 0.037 & 0.105 & 0.537 \\
LLama3.2-Vision (11B) \cite{llama} & 0.246 & 0.075 & 0.238 & 0.521 & 0.239 & 0.079 & 0.325 & 0.471 & 0.532 & 0.188 & 0.349 & 0.466 & 0.258 & 0.051 & 0.239 & 0.475 & 0.268 & 0.321 & 0.237 & 0.595 & 0.538 & 0.127 & 0.383 & 0.537 \\
MiniCPM-V2.6 (8B) \cite{minicpm} & 0.282 & 0.060 & 0.368 & 0.651 & 0.518 & 0.312 & 0.508 & 0.576 & 0.716 & 0.320 & 0.416 & 0.521 & 0.276 & 0.126 & 0.312 & 0.576 & 0.584 & 0.280 & 0.312 & 0.608 & 0.534 & 0.178 & 0.422 & 0.593 \\
InternVL2 (8B) \cite{internvl} & 0.320 & 0.604 & 0.008 & 0.798 & 0.395 & 0.419 & 0.180 & 0.727 & 0.745 & 0.371 & 0.384 & 0.395 & 0.314 & 0.439 & 0.152 & 0.716 & 0.526 & 0.375 & 0.370 & 0.630 & 0.633 & 0.414 & 0.230 & 0.671 \\
InternVL2.5 (8B) \cite{internvl2} & 0.304 & 0.611 & 0.007 & 0.803 & 0.403 & 0.484 & 0.177 & 0.731 & 0.750 & 0.377 & 0.368 & 0.399 & 0.325 & 0.444 & 0.148 & 0.716 & 0.533 & 0.386 & 0.366 & 0.627 & 0.691 & 0.421 & 0.225 & 0.670 \\
Qwen2-VL (7B) \cite{qwenvl} & 0.424 & 0.008 & 0.522 & 0.534 & 0.637 & 0.166 & 0.687 & 0.546 & 0.734 & 0.253 & 0.436 & 0.508 & 0.419 & 0.023 & 0.562 & 0.495 & 0.630 & 0.398 & 0.512 & 0.565 & 0.679 & 0.194 & 0.548 & 0.531 \\
Qwen2.5-VL (7B) \cite{qwenvl2} & 0.537 & 0.375 & 0.674 & 0.790 & 0.660 & 0.473 & 0.769 & 0.803 & 0.551 & 0.222 & 0.613 & 0.786 & 0.440 & 0.265 & 0.669 & 0.787 & 0.481 & 0.266 & 0.754 & 0.693 & 0.682 & 0.251 & 0.678 & 0.738 \\
DeepSeekVL (7B) \cite{deepseekvl} & 0.296 & 0.369 & 0.221 & 0.454 & 0.205 & 0.001 & 0.204 & 0.399 & 0.720 & 0.007 & 0.229 & 0.303 & 0.269 & 0.169 & 0.179 & 0.439 & 0.425 & 0.271 & 0.045 & 0.424 & 0.679 & 0.173 & 0.237 & 0.431 \\
DeepSeekVL2 (small) \cite{deepseekvl2} &0.257 & 0.316 & 0.265 & 0.454 & 0.200 & 0.066 & 0.308 & 0.399 & 0.688 & 0.013 & 0.282 & 0.303 & 0.257 & 0.146 & 0.210 & 0.439 & 0.391 & 0.309 & 0.109 & 0.424 & 0.668 & 0.176 & 0.301 & 0.431 \\
\hline
\noalign{\vspace{1pt}}

\raisebox{0.5ex}{\scriptsize \ding{91}}LLaVA-NeXT (8B) \cite{llava-n} & 0.730 & 0.832 & 0.826 & 0.887 & 0.827 & 0.886 & 0.864 & 0.870 & 0.860 & \mbluebf{0.721} & 0.841 & \mbluebf{0.845} & 0.717 & 0.819 & 0.790 & \mbluebf{0.864} & 0.761 & 0.802 & 0.850 & 0.848 & 0.860 & 0.813 & 0.852 & 0.856 \\
\raisebox{0.5ex}{\scriptsize \ding{91}}DeepSeekVL2 (small) \cite{deepseekvl2} & 0.749 & \mbluebf{0.869} & \mbluebf{0.855} & \mbluebf{0.895} & \mbluebf{0.859} & \mbluebf{0.893} & 0.890 & 0.857 & 0.852 & 0.716 & \mbluebf{0.885} & 0.841 & \mbluebf{0.764} & \mbluebf{0.829} & \mbluebf{0.820} & 0.861 & 0.759 & \mbluebf{0.828} & 0.883 & 0.859 & 0.867 & \mbluebf{0.828} & 0.878 & 0.860 \\
\raisebox{0.5ex}{\scriptsize \ding{91}}InternVL2.5 (8B) \cite{internvl2} & \mbluebf{0.759} & 0.846 & 0.853 & 0.895 & 0.857 & 0.866 & \mbluebf{0.903} & \mbluebf{0.874} & \mbluebf{0.874} & 0.653 & 0.883 & 0.811 & 0.751 & 0.825 & 0.816 & 0.859 & \mbluebf{0.819} & 0.825 & \mbluebf{0.894} & \mbluebf{0.869} & \mbluebf{0.884} & 0.821 & \mbluebf{0.884} & 0.864 \\
\rowcolor{gray!20}  
\raisebox{0.5ex}{\scriptsize \ding{91}}LMM4Edit (Ours) 
& \mredbf{0.807} & \mredbf{0.879} & \mredbf{0.875} & \mredbf{0.924} 
& \mredbf{0.897} & \mredbf{0.922} & \mredbf{0.923} & \mredbf{0.891} 
& \mredbf{0.910} & \mredbf{0.798} & \mredbf{0.892} & \mredbf{0.849} 
& \mredbf{0.791} & \mredbf{0.872} & \mredbf{0.856} & \mredbf{0.887} 
& \mredbf{0.874} & \mredbf{0.887} & \mredbf{0.914} & \mredbf{0.869} 
& \mredbf{0.914} & \mredbf{0.883} & \mredbf{0.905} & \mredbf{0.878} \\

 \noalign{\vspace{-1.5pt}}
\bottomrule
\end{tabular}}
\end{table*}
\section{LMM4Edit}
In this section, we present \textbf{LMM4Edit}, the first \textbf{all-in-one} TIE evaluation model to answer task-specific questions, and give fine-grained perceptual quality, editing alignment and attribute preservation scores aligned with human perception.
\subsection{Model Design}
\textbf{Overall Architecture.} The overall framework of LMM4Edit is shown in Figure~\ref{model}. It takes the edited image, source image, and prompt (editing instruction \& evaluation dimension) as input to predict fine-grained perceptual quality, editing alignment, and attribute scores and answer task-specific questions. LMM4Edit begins by extracting visual and text features from the images and user prompts, respectively. A weight-frozen vision encoder extracts image features, which are then projected into the language space via a projector, generating visual tokens $T_e$ for edited image and $T_s$ for source image. For text feature extraction, a tokenizer encodes the user prompt into prompt tokens $T_p$. The concatenated tokens $T_e$, $T_s$ and $T_p$ are fed into a pre-trained LLM. The output last hidden states are decoded through a text decoder in the first training stage or a quality score decoder in the second training stage.

\textbf{Visual Encoding.} The image encoder $E_{i}$ is based on the pre-trained vision transformer, 
CLIP-ViT-bigG \cite{ViT}. To align the extracted features with the input space of the LLM, a trainable projector $P_{i}$ with two multi-layer perception (MLP) layers is applied. This projects the image features into a language space, generating the visual feature tokens. For the edited image $I_{e}$ and its source image $I_{s}$ is input. The process can be formulated as: 
\begin{equation}
    T_e = P_{i}(E_{i}(I_{e})) \quad
    T_s = P_{i}(E_{i}(I_{s}))
\end{equation}
where $T_e$ and $T_s$ are the visual tokens for edited image and source image, respectively. 

\textbf{Feature Fusion via the LLM.} We combine the image editing instruction into the user prompt, which is first encoded into text tokens $T_p$ using a tokenizer. These text tokens $T_p$ are then concatenated with the well-aligned visual tokens $T_e$ and $T_s$ to form the input to the LLM. Specifically, the pre-trained QwenLM2 \cite{qwenvl} is used to combine the visual and text tokens for multimodal learning.

\textbf{Adaptive Decoding.}
To enable the model to generate fine-grained scores, adaptive decoding is utilized. The last hidden states output by the LLM are decoded by text decoder firstly. Once the model is capable of generating responses in the desired format and content, the hidden state representing the token just before the score is then passed to a quality score decoder. This decoder, consisting of two MLPs, is employed in the second training stage to yield a more precise quality score.

\begin{table*}[t]
\belowrulesep=0pt
\aboverulesep=0pt
\centering
\renewcommand{\arraystretch}{0.85}
\caption{Comparisons of the alignment between different evaluation methods and human perception in evaluating TIE models. The best results are highlighted in \mredbf{red}, and the second-best results are highlighted in \mbluebf{blue}. \raisebox{0.5ex}{\scriptsize \ding{91}} denotes fine-tuned models.} 
\resizebox{\textwidth}{!}{
\begin{tabular}{l||c:ccc| c:ccc| c:ccc| c:ccc| c:c |c:c}
\toprule
\noalign{\vspace{1.5pt}}
Dimensions & \multicolumn{4}{c}{Perceptual Quality} & \multicolumn{4}{c}{Editing Alignment} & \multicolumn{4}{c}{Attribute Preservation}& \multicolumn{4}{c}{Task-specific Accuracy ($\%$)}& \multicolumn{2}{c}{Overall Rank} &\multicolumn{2}{c}{Acc Rank} \\
\cmidrule(lr){2-5}
\cmidrule(lr){6-9}
\cmidrule(lr){10-13}
\cmidrule(lr){14-17}
\cmidrule(lr){18-19}
\cmidrule(lr){20-21}
\noalign{\vspace{1.5pt}}
Models/Metrics & Human &\cellcolor{gray!20} Ours\raisebox{0.5ex}{\scriptsize \ding{91}} & Q-Align\raisebox{0.5ex}{\scriptsize \ding{91}} & MANIQA\raisebox{0.5ex}{\scriptsize \ding{91}} & Human &\cellcolor{gray!20} Ours\raisebox{0.5ex}{\scriptsize \ding{91}}  &Q-Align\raisebox{0.5ex}{\scriptsize \ding{91}}&PickScore &Human&\cellcolor{gray!20} Ours\raisebox{0.5ex}{\scriptsize \ding{91}}&AHIQ\raisebox{0.5ex}{\scriptsize \ding{91}}&LPIPS&Human&\cellcolor{gray!20} Ours\raisebox{0.5ex}{\scriptsize \ding{91}} & Intern2.5\raisebox{0.5ex}{\scriptsize \ding{91}}&Qwen2.5&Human&\cellcolor{gray!20}Ours&Human&\cellcolor{gray!20}Ours\\

\hline
\noalign{\vspace{1.5pt}}
FlowEdit (SD3) \cite{Flowedit} & 54.70 & \cellcolor{gray!20}52.64 & 57.95 & 65.82 & 52.61 & \cellcolor{gray!20}52.65 & 55.14 & 63.22 & 55.57 & \cellcolor{gray!20}54.40 & 62.50 & 90.26 & 72.22 & \cellcolor{gray!20}74.54 & 75.00 & 41.67& 1 & \cellcolor{gray!20}1 & 1 & \cellcolor{gray!20}1\\
PnP \cite{PnP} & 52.63 & 
\cellcolor{gray!20}51.20 & 55.56 & 65.57 & 55.67 & \cellcolor{gray!20}56.17 & 58.79 & 58.91 & 51.11 & \cellcolor{gray!20}50.21 & 55.59 & 85.82 & 68.06 & \cellcolor{gray!20}70.83 & 66.20 & 44.91& 2 & \cellcolor{gray!20}2 & 2 & \cellcolor{gray!20}2\\
RFSE \cite{RFSE}& 55.91 & 
\cellcolor{gray!20}53.23 & 58.70 & 68.57 & 56.35 & \cellcolor{gray!20}56.73 & 59.36 & 64.35 & 46.74 & \cellcolor{gray!20}45.47 & 47.77 & 82.70 & 67.13 & \cellcolor{gray!20}64.35 & 60.65 & 40.28 & 3 & \cellcolor{gray!20}3 & 3 & \cellcolor{gray!20}4\\
CDS \cite{CDS} & 54.28 & 
\cellcolor{gray!20}51.81 & 53.21 & 61.67 & 46.01 & \cellcolor{gray!20}46.07 & 49.06 & 50.80 & 63.05 & \cellcolor{gray!20}61.51 & 76.42 & 96.73 & 64.81 & \cellcolor{gray!20}66.67 & 59.72 & 25.93 & 4 & \cellcolor{gray!20}4 & 4 & \cellcolor{gray!20}3\\
InfEdit \cite{InfEdit} & 51.27 &
\cellcolor{gray!20}49.44 & 51.49 & 59.76 & 54.96 & \cellcolor{gray!20}54.97 & 56.36 & 55.49 & 52.02 & \cellcolor{gray!20}50.63 & 57.62 & 87.84 & 49.54 & \cellcolor{gray!20}53.70 & 45.37 & 47.69 & 5 & \cellcolor{gray!20}5 & 5 & \cellcolor{gray!20}6\\
FlowEdit (FLUX) \cite{Flowedit} & 50.56 &
\cellcolor{gray!20}50.04 & 54.10 & 62.47 & 52.08 & \cellcolor{gray!20}53.25 & 52.96 & 60.80 & 51.43 & \cellcolor{gray!20}50.52 & 58.95 & 89.13 & 49.54 & \cellcolor{gray!20}52.31 & 44.44 & 43.52 & 6 & \cellcolor{gray!20}6 & 6 & \cellcolor{gray!20}7\\
EDICT\cite{EDICT} & 50.63 &
\cellcolor{gray!20}49.10 & 51.02 & 60.59 & 47.66 & \cellcolor{gray!20}48.80 & 51.44 & 47.66 & 57.05 & \cellcolor{gray!20}55.83 & 69.44 & 93.62 & 49.07 & \cellcolor{gray!20}54.63 & 43.98 &33.33& 7 & \cellcolor{gray!20}8 & 7 & \cellcolor{gray!20}5\\
Any2Pix \cite{instructany2pix}& 54.04 &
\cellcolor{gray!20}52.81 & 58.90 & 67.69 & 57.55 & \cellcolor{gray!20}58.38 & 61.39 & 57.14 & 41.15 & \cellcolor{gray!20}41.37 & 36.78 & 76.51 & 44.91 & \cellcolor{gray!20}45.83 & 43.98 &34.26 & 8 & \cellcolor{gray!20}7 & 8 & \cellcolor{gray!20}9\\
Magicbrush \cite{Magicbrush} & 49.18 &
\cellcolor{gray!20}48.57 & 50.69 & 59.23 & 49.73 & \cellcolor{gray!20}50.02 & 48.21 & 45.07 & 52.48 & \cellcolor{gray!20}51.89 & 60.13 & 86.98 & 41.20 & \cellcolor{gray!20}46.30 & 45.37 & 32.87& 9 & \cellcolor{gray!20}9 & 9 & \cellcolor{gray!20}8\\
ZONE \cite{zone}& 49.67 &
\cellcolor{gray!20}48.87 & 50.37 & 60.33 & 44.95 & \cellcolor{gray!20}45.15 & 46.02 & 47.72 & 58.77 & \cellcolor{gray!20}58.67 & 73.68 & 95.06 & 37.50 & \cellcolor{gray!20}40.28 & 35.65 &18.98 & 10 & \cellcolor{gray!20}10 & 10 & \cellcolor{gray!20}10\\
ReNoise \cite{renoise} & 50.20 & 
\cellcolor{gray!20}48.78 & 51.31 & 60.58 & 51.19 & \cellcolor{gray!20}51.84 & 53.12 & 54.80 & 48.16 & \cellcolor{gray!20}47.18 & 53.22 & 85.91 & 32.41 & \cellcolor{gray!20}29.63 & 25.46 & 28.70 & 11 & \cellcolor{gray!20}11 & 11 & \cellcolor{gray!20}13\\
IP2P \cite{ip2p} & 48.28 & 
\cellcolor{gray!20}47.44 & 48.47 & 56.08 & 46.73 & \cellcolor{gray!20}47.34 & 47.97 & 45.20 & 52.72 & \cellcolor{gray!20}51.79 & 60.76 & 87.67 & 31.94 & \cellcolor{gray!20}34.26 & 30.56 & 24.54 & 12 & \cellcolor{gray!20}12 & 12 & \cellcolor{gray!20}11\\
ACE++ \cite{ACE} & 50.83 &
\cellcolor{gray!20}49.25 & 51.45 & 60.48 & 46.39 & \cellcolor{gray!20}46.73 & 48.19 & 43.90 & 44.34 & \cellcolor{gray!20}43.97 & 37.33 & 75.28 & 30.56 & \cellcolor{gray!20}30.09 & 24.54 & 18.52 & 13 & \cellcolor{gray!20}13 & 13 & \cellcolor{gray!20}12\\
HQEdit \cite{HQ} & 46.81 &
\cellcolor{gray!20}46.61 & 47.30 & 57.48 & 48.23 & \cellcolor{gray!20}50.05 & 49.74 & 43.90 & 39.21 & \cellcolor{gray!20}38.78 & 34.21 & 73.77 & 26.85 & \cellcolor{gray!20}23.15 & 19.91 & 16.67 & 14 & \cellcolor{gray!20}14 & 14 & \cellcolor{gray!20}15\\
MasaCtrl \cite{Masactrl} & 44.13 &
\cellcolor{gray!20}43.88 & 43.75 & 52.00 & 45.07 & \cellcolor{gray!20}44.74 & 44.53 & 47.91 & 44.07 & \cellcolor{gray!20}44.30 & 49.19 & 84.68 & 26.39 & \cellcolor{gray!20}23.15 & 25.00 & 22.22 & 15 & \cellcolor{gray!20}15 & 15 & \cellcolor{gray!20}16\\
DDPM \cite{DDPM} & 44.72 &
\cellcolor{gray!20}43.73 & 43.73 & 51.96 & 40.83 & \cellcolor{gray!20}41.08 & 42.24 & 30.68 & 49.59 & \cellcolor{gray!20}48.19 & 61.82 & 91.29 & 25.93 & \cellcolor{gray!20}26.39 & 20.83 & 10.19 & 16 & \cellcolor{gray!20}16 & 16 & \cellcolor{gray!20}14\\
Text2LIVE \cite{Text2live}& 37.50 &
\cellcolor{gray!20}39.13 & 37.00 & 40.18 & 44.55 & \cellcolor{gray!20}43.07 & 44.32 & 45.97 & 46.46 & \cellcolor{gray!20}47.38 & 57.80 & 88.94 & 11.11 & \cellcolor{gray!20}9.72 & 10.65 & 32.41 & 17 & \cellcolor{gray!20}17 & 17 & \cellcolor{gray!20}17\\

\hline
\noalign{\vspace{1.5pt}}
SRCC to human $\uparrow$&&\cellcolor{gray!20}\mredbf{0.973}&
\mbluebf{0.941}&0.914&&\cellcolor{gray!20}\mredbf{0.993}&\mbluebf{0.963}&0.722&&\cellcolor{gray!20}\mredbf{0.988}&\mbluebf{0.919}&0.838&&\cellcolor{gray!20}\mredbf{0.972}&\mbluebf{0.964}& 0.803&&\cellcolor{gray!20}\mredbf{0.998}&&\cellcolor{gray!20}\mredbf{0.973}\\
RMSE to human $\downarrow$&&\cellcolor{gray!20}\mredbf{1.480}&\mbluebf{2.040}&10.05&&\cellcolor{gray!20}\mredbf{0.785}&\mbluebf{2.210}&5.250&&\cellcolor{gray!20}\mredbf{0.965}&\mbluebf{8.600}&36.48&&\cellcolor{gray!20}\mredbf{3.010}&\mbluebf{4.420}&16.44&&\cellcolor{gray!20}\mredbf{0.343}&&\cellcolor{gray!20}\mredbf{1.138}\\
\bottomrule
\end{tabular}
}
\label{compare_TIE}
\end{table*}

\begin{table*}[tb]
\renewcommand{\arraystretch}{0.85}
\centering
\caption{Ablation study on the different backbones, projector training and LoRA tuning strategy.}
\label{ablation}
\fontsize{2.8}{3}\selectfont
\setlength{\arrayrulewidth}{0.15pt}  
\setlength{\heavyrulewidth}{0.08em}  
\setlength{\cmidrulewidth}{0.15pt}    
 \resizebox{1\textwidth}{!}{
\begin{tabular}{lcccc|ccc|ccc|ccc|c}
\toprule
 \noalign{\vspace{-1.3pt}}
\multicolumn{5}{c}{Backbone$\&$Strategy} & \multicolumn{3}{c}{Perceptual Quality} & \multicolumn{3}{c}{Editing Alignment} & \multicolumn{3}{c}{Attribute Preservation} &QA\\
 \noalign{\vspace{-1.0pt}}
\cmidrule(lr){1-5} \cmidrule(lr){6-8} \cmidrule(lr){9-11} \cmidrule(lr){12-14} \cmidrule(lr){15-15}
 \noalign{\vspace{-1.5pt}}
Backbone&Train Projector &  LoRA(vision) & LoRA(llm) & LoRA Type&  SRCC & KRCC &PLCC & SRCC  & KRCC & PLCC& SRCC & KRCC &PLCC &Acc\\ 
\hline
\noalign{\vspace{1pt}}

Qwen2.5-VL \cite{qwenvl2} & & \checkmark &   &AdaLoRA& 0.8322 & 0.6963 &0.8471 & 0.8105&0.6782 &0.8247 & 0.8652&0.7022 &0.8741& 84.3\\
Qwen2.5-VL \cite{qwenvl2}& &  & \checkmark &AdaLoRA & 0.8018 & 0.6792 &0.8163 & 0.7946&0.6294 &0.8062 & 0.8273&0.6738 &0.8343& 83.5\\
Qwen2.5-VL \cite{qwenvl2}& & \checkmark &  \checkmark &AdaLoRA & 0.8653 & 0.7291 &0.8799 & 0.8622&0.6844 &0.8781 & 0.8811&0.7502 &0.8971& 85.7\\
\rowcolor{gray!20}  
Qwen2.5-VL \cite{qwenvl2}&\checkmark & \checkmark &  \checkmark &AdaLoRA & \textbf{0.9136} & \textbf{0.7432} &\textbf{0.9189} & \textbf{0.8830}&\textbf{0.7080} &\textbf{0.8898} & \textbf{0.9048}&\textbf{0.7837} &\textbf{0.9176}&\textbf{87.8}\\
Qwen2.5-VL \cite{qwenvl2}&\checkmark & \checkmark &  \checkmark &LoRA & 0.8974 & 0.7312 &0.9033 & 0.8711&0.6819 &0.8815 & 0.8902&0.7746 &0.8986& 87.2\\
InternVL2.5 \cite{internvl2}&\checkmark & \checkmark &  \checkmark &AdaLoRA & 0.8836 & 0.7223 &0.8861 & 0.8207&0.6809 &0.8387 & 0.8841&0.7571 &0.8949& 86.4\\
DeepSeekVL2 \cite{deepseekvl2} &\checkmark & \checkmark &  \checkmark &AdaLoRA & 0.8674 & 0.7303 &0.8665 & 0.8280&0.6893&0.8390&0.8778&0.7500&0.8819&86.0 \\
LLaVA-NeXT \cite{llava}&\checkmark & \checkmark &  \checkmark &AdaLoRA & 0.8601 & 0.7251 &0.8657 & 0.8125&0.6714 &0.8241 & 0.8517&0.7206 &0.8494& 85.6\\

 \noalign{\vspace{-1.5pt}}
\bottomrule
\end{tabular}}
\end{table*}
\setlength{\textfloatsep}{1pt plus 0.2pt minus 0.5pt}
\setlength{\dbltextfloatsep}{1pt plus 0.2pt minus 0.5pt}
\setlength{\dblfloatsep}{1pt plus 0.2pt minus 0.5pt}
\setlength{\intextsep}{1pt plus 0.2pt minus 0.5pt}
\setlength{\abovecaptionskip}{1pt plus 0.2pt minus 0.5pt}
\setlength{\belowcaptionskip}{1pt plus 0.2pt minus 0.5pt}
\setlength{\abovedisplayskip}{1.5pt}
\setlength{\belowdisplayskip}{1.5pt}
\setlength{\topsep}{1.5pt}
\titlespacing*{\section}{0pt}{1pt plus 0.2pt minus 0.5pt}{1pt plus 0.2pt minus 0.5pt} 
\titlespacing*{\subsection}{0pt}{1pt plus 0.2pt minus 0.5pt}{1pt plus 0.2pt minus 0.5pt} 
\subsection{Fine-tuning Techniques}\indent

\textbf{Instruction Tuning.} Achieving an \textbf{all-in-one} image quality assessment model is important for enabling multi-dimensional quality evaluation within a single model. We employ the instruction-tuning strategy \cite{Ituning} to train the model for task-specific question answering and quality score prediction. As shown in Figure~\ref{model}, our user prompt includes clear and explicit problem descriptions tailored to different TIE evaluation tasks, allowing LMM4Edit to accurately respond to specific requirements.

\textbf{AdaLoRA Adaptation.}
To enhance the performance, we employ the AdaLoRA technique \cite{adalora} for efficient model adaptation in pre-trained LMMs. Unlike standard LoRA, which applies a fixed low-rank decomposition to model updates, AdaLoRA introduces an adaptive mechanism to dynamically allocate rank across different layers based on their importance.  
AdaLoRA extends this by dynamically adjusting the rank across different layers, progressively redistributing parameters from less important layers to more critical ones during training. This adaptive strategy ensures that computational resources are allocated efficiently, leading to improved model adaptation. By integrating AdaLoRA, LMM4Edit can better adapt to the TIE evaluation while maintaining parameter efficiency.  

\textbf{Two-stage Training.}
We train LMM4Edit in two stages. In the first stage, we use cross-entropy loss, with the label being text sentences. The goal of this stage is to train the model to generate text in the desired format, along with answering task-specific questions and predicting rough quality scores. However, relying solely on text training does not yield an accurate score result. Therefore, in the second stage, we use Mean Squared Error (MSE) loss, with the label being the quality score number. The objective of this stage is to refine the ability of LMM4Edit to produce accurate perceptual quality, editing alignment and attribute preservation scores.
\section{Experiments}
In this section, we evaluate the performance of our LMM4Edit model through extensive experiments.
\subsection{Experiment Setup}
To evaluate the correlation between the predicted scores and the ground-truth MOSs, we use three evaluation metrics, including Spearman Rank Correlation Coefficient (SRCC), Pearson Linear Correlation Coefficient (PLCC), and Kendall’s Rank Correlation Coefficient (KRCC). For visual question answering, we adopt accuracy as the metric. 

We apply numerous metrics for comparison. First, traditional handcrafted IQA metrics are used directly for evaluation. For vision-language and LLM-based models, we use pre-trained weights for inference. Three LLM-based models are also fine-tuned using the same approach as the backbone of our model. For learning-based models, we use the same training and testing split (4:1) as in previous literature. The models are implemented with PyTorch and trained on a 40GB NVIDIA RTX A100 GPU with a batch size of 1 and gradient accumulation steps of 16. The initial learning rate is set to 1e-4 and is decreased using the cosine annealing strategy. During pre-training, the number of training epochs is set to 1, and for fine-tuning, it is set to 3.
\subsection{Evaluation on EBench-18K Dataset}
Table~\ref{performances} presents the performance of our LMM4Edit in comparison with traditional and deep learning-based FR IQA and NR IQA methods, vision-language methods, and LMM-based models. Traditional IQA methods perform poorly, as their features are primarily designed to assess conventional image distortions and are ineffective in capturing structural distortions and evaluating editing alignment. Deep learning-based IQA methods achieve better results in assessing perceptual quality but still struggle with evaluating editing alignment. We also find vision-language pretraining models also show limited effectiveness, as they are designed to evaluate text-image alignment, which differs from the editing alignment. Although fine-tuned LMM-based models perform well in TIE evaluation generally, their ability to assess editing alignment remains inadequate. Unlike these methods, our LMM4Edit achieves outstanding performance in aligning with human perception across perceptual quality, editing alignment, and attribute preservation.
Figure~\ref{LMM} and Table~\ref{comparison_task} further compare the performance of LMM-based models across the 21 editing tasks on our EBench-18K in detail. The zero-shot results of LMMs struggle to evaluate complex high-level editing tasks, such as action and size, as well as low-level tasks that require detailed capturing, such as deraining and super-resolution. However, when fine-tuned using our proposed methods, their performance improves significantly. Our model achieves superior performance in both score prediction and yes/no question answering, establishing it as a more comprehensive solution for evaluating TIE.
\begin{figure}[t]
\vspace{5pt}
  \includegraphics[width=0.38\textwidth]{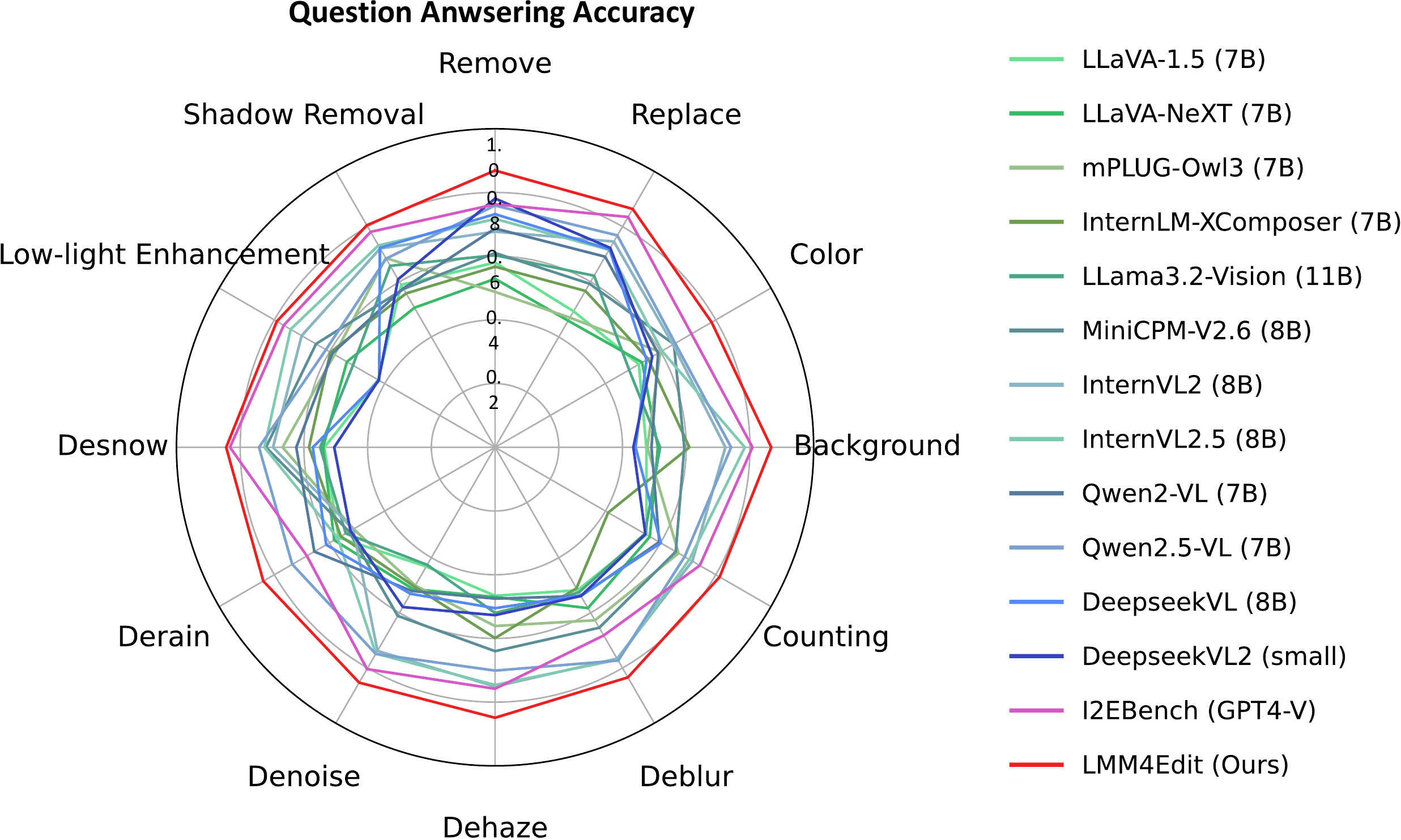}
  \caption{Zero-shot cross-dataset performance comparison on I2EBench \cite{I2EBench} of our model and advanced LMMs in terms of QA accuracy.}
  \label{zero-shot}
\end{figure}

\titlespacing*{\section}{0pt}{1.5pt plus 0.5pt minus 0.5pt}{1.5pt plus 0.5pt minus 0.5pt} 
\titlespacing*{\subsection}{0pt}{1.5pt plus 0.5pt minus 0.5pt}{1.5pt plus 0.5pt minus 0.5pt} 
\setlength{\abovedisplayskip}{2.5pt}
\setlength{\belowdisplayskip}{2.5pt}

\subsection{Evaluation on TIE Model Performance}
We further compare the alignment between different metric results and human annotations in evaluating TIE model performance, as shown in Table~\ref{compare_TIE}. The overall rank is derived from a combination of the perceptual quality score $Score_{q}$, editing alignment score $Score_{e}$ and attribute preservation score $Score_{p}$ through the following equation: 
\begin{equation}
Score_{all}=Score_{q}^{0.3}*Score_{e}^{0.4}*Score_{p}^{0.3}   
\end{equation}
The editing alignment score is given a higher weight to emphasize the importance of aligning with the editing expectation.
Our model achieves the highest SRCC with human ratings and the lowest relative Root Mean Square Error (RMSE) in score differences. This demonstrates our model's ability to accurately assess and rank the performance of TIE models.
\subsection{Ablation Study}
To validate the effectiveness of the different modules in LMM4Edit, we conduct comprehensive ablation studies, with the results summarized in Table~\ref{ablation}. Rows 1 to 3 of Table~\ref{ablation} show that applying LoRA to both the vision model and the LLM yields the best performance for our model. The projector in LMM4Edit maps vision features into the language space, and as shown in row 4, training its weights further enhances performance. AdaLoRA dynamically allocates trainable parameters, improving efficiency and adaptability compared to standard LoRA, as demonstrated in row 5. Rows 6 to 8 present the performance of other LMM backbones with the same parameter scale, where Qwen2.5-VL achieves the best results.
\subsection{Zero-shot Cross-dataset Evaluation}
Among existing TIE benchmarks, only I2EBench \cite{I2EBench} provides human annotations in the form of QA pairs to assess editing accuracy, and GPT-4V \cite{gpt4} is used for evaluation. We compare the zero-shot performance of LMM4Edit with other leading LMMs on I2EBench. As shown in Figure~\ref{zero-shot}, our model achieves the best zero-shot performance, demonstrating its strong generalization ability.
\section{Conclusion}
In this paper, we introduce EBench-18K, the first large-scale dataset for evaluating TIE models from multiple dimensions, consisting of 18K+ images and 1M+ annotations. Based on EBench-18K, we propose LMM4Edit, the first all-in-one LMM-based metric for evaluating TIE models by predicting fine-grained perceptual quality, editing alignment, attribute preservation scores, and answering task-specific questions. Through extensive experiments, we demonstrate that LMM4Edit outperforms all existing methods, exhibiting better alignment with human preference and superior generalization ability.


\clearpage
\newpage
\bibliographystyle{ACM-Reference-Format}
\bibliography{sample-base}


\begin{thebibliography}{93}


\ifx \showCODEN    \undefined \def \showCODEN     #1{\unskip}     \fi
\ifx \showISBNx    \undefined \def \showISBNx     #1{\unskip}     \fi
\ifx \showISBNxiii \undefined \def \showISBNxiii  #1{\unskip}     \fi
\ifx \showISSN     \undefined \def \showISSN      #1{\unskip}     \fi
\ifx \showLCCN     \undefined \def \showLCCN      #1{\unskip}     \fi
\ifx \shownote     \undefined \def \shownote      #1{#1}          \fi
\ifx \showarticletitle \undefined \def \showarticletitle #1{#1}   \fi
\ifx \showURL      \undefined \def \showURL       {\relax}        \fi
\providecommand\bibfield[2]{#2}
\providecommand\bibinfo[2]{#2}
\providecommand\natexlab[1]{#1}
\providecommand\showeprint[2][]{arXiv:#2}

\bibitem[Abdelhamed et~al\mbox{.}(2018)]%
        {SSID}
\bibfield{author}{\bibinfo{person}{Abdelrahman Abdelhamed}, \bibinfo{person}{Stephen Lin}, {and} \bibinfo{person}{Michael~S Brown}.} \bibinfo{year}{2018}\natexlab{}.
\newblock \showarticletitle{A high-quality denoising dataset for smartphone cameras}. In \bibinfo{booktitle}{\emph{Proceedings of the IEEE/CVF Conference on Computer Vision and Pattern Recognition (CVPR)}}. \bibinfo{pages}{1692--1700}.
\newblock


\bibitem[{AI Meta}(2024)]%
        {llama}
\bibfield{author}{\bibinfo{person}{{AI Meta}}.} \bibinfo{year}{2024}\natexlab{}.
\newblock \showarticletitle{Llama 3.2: Revolutionizing Edge AI and Vision with Open, Customizable Models}.
\newblock \bibinfo{journal}{\emph{Meta AI Blog}} (\bibinfo{date}{20 December} \bibinfo{year}{2024}).
\newblock


\bibitem[Bai et~al\mbox{.}(2025)]%
        {qwenvl2}
\bibfield{author}{\bibinfo{person}{Shuai Bai}, \bibinfo{person}{Keqin Chen}, \bibinfo{person}{Xuejing Liu}, \bibinfo{person}{Jialin Wang}, \bibinfo{person}{Wenbin Ge}, \bibinfo{person}{Sibo Song}, {et~al\mbox{.}}} \bibinfo{year}{2025}\natexlab{}.
\newblock \showarticletitle{Qwen2.5-VL Technical Report}.
\newblock \bibinfo{journal}{\emph{arXiv preprint arXiv:2502.13923}} (\bibinfo{year}{2025}).
\newblock


\bibitem[Bar-Tal et~al\mbox{.}(2022)]%
        {Text2live}
\bibfield{author}{\bibinfo{person}{Omer Bar-Tal}, \bibinfo{person}{Dolev Ofri-Amar}, \bibinfo{person}{Rafail Fridman}, \bibinfo{person}{Yoni Kasten}, {and} \bibinfo{person}{Tali Dekel}.} \bibinfo{year}{2022}\natexlab{}.
\newblock \showarticletitle{Text2live: Text-driven layered image and video editing}. In \bibinfo{booktitle}{\emph{Proceedings of the European Conference on Computer Vision (ECCV)}}. \bibinfo{pages}{707--723}.
\newblock


\bibitem[Basu et~al\mbox{.}(2023)]%
        {EditVal}
\bibfield{author}{\bibinfo{person}{Samyadeep Basu}, \bibinfo{person}{Mehrdad Saberi}, \bibinfo{person}{Shweta Bhardwaj}, \bibinfo{person}{Atoosa~Malemir Chegini}, \bibinfo{person}{Daniela Massiceti}, \bibinfo{person}{Maziar Sanjabi}, {et~al\mbox{.}}} \bibinfo{year}{2023}\natexlab{}.
\newblock \showarticletitle{EditVal: Benchmarking Diffusion Based Text-Guided Image Editing Methods}.
\newblock \bibinfo{journal}{\emph{arXiv preprint arXiv:2310.02426}} (\bibinfo{year}{2023}).
\newblock


\bibitem[Bosse et~al\mbox{.}(2017)]%
        {Wa}
\bibfield{author}{\bibinfo{person}{S. Bosse}, \bibinfo{person}{D. Maniry}, \bibinfo{person}{K.-R. Muller}, \bibinfo{person}{T. Wiegand}, {and} \bibinfo{person}{W. Samek}.} \bibinfo{year}{2017}\natexlab{}.
\newblock \showarticletitle{Deep neural networks for no-reference and full-reference image quality assessment}.
\newblock \bibinfo{journal}{\emph{IEEE Transactions on Image Processing (TIP)}} \bibinfo{volume}{27}, \bibinfo{number}{1} (\bibinfo{year}{2017}), \bibinfo{pages}{206--219}.
\newblock


\bibitem[Brooks et~al\mbox{.}(2023)]%
        {ip2p}
\bibfield{author}{\bibinfo{person}{Tim Brooks}, \bibinfo{person}{Aleksander Holynski}, {and} \bibinfo{person}{Alexei~A Efros}.} \bibinfo{year}{2023}\natexlab{}.
\newblock \showarticletitle{Instructpix2pix: Learning to follow image editing instructions}. In \bibinfo{booktitle}{\emph{Proceedings of the IEEE/CVF Conference on Computer Vision and Pattern Recognition (CVPR)}}. \bibinfo{pages}{18392--18402}.
\newblock


\bibitem[Cao et~al\mbox{.}(2023)]%
        {Masactrl}
\bibfield{author}{\bibinfo{person}{Mingdeng Cao}, \bibinfo{person}{Xintao Wang}, \bibinfo{person}{Zhongang Qi}, \bibinfo{person}{Ying Shan}, \bibinfo{person}{Xiaohu Qie}, {and} \bibinfo{person}{Yinqiang Zheng}.} \bibinfo{year}{2023}\natexlab{}.
\newblock \showarticletitle{Masactrl: Tuning-free mutual self-attention control for consistent image synthesis and editing}. In \bibinfo{booktitle}{\emph{Proceedings of the IEEE/CVF international conference on computer vision (CVPR)}}. \bibinfo{pages}{22560--22570}.
\newblock


\bibitem[Chen et~al\mbox{.}(2024a)]%
        {TOPIQ}
\bibfield{author}{\bibinfo{person}{Chaofeng Chen}, \bibinfo{person}{Jiadi Mo}, \bibinfo{person}{Jingwen Hou}, \bibinfo{person}{Haoning Wu}, \bibinfo{person}{Liang Liao}, \bibinfo{person}{Wenxiu Sun}, \bibinfo{person}{Qiong Yan}, {and} \bibinfo{person}{Weisi Lin}.} \bibinfo{year}{2024}\natexlab{a}.
\newblock \showarticletitle{TOPIQ: A Top-Down Approach From Semantics to Distortions for Image Quality Assessment}.
\newblock \bibinfo{journal}{\emph{IEEE Transactions on Image Processing (TIP)}}  \bibinfo{volume}{33} (\bibinfo{year}{2024}), \bibinfo{pages}{2404--2418}.
\newblock


\bibitem[Chen et~al\mbox{.}(2021)]%
        {CSD}
\bibfield{author}{\bibinfo{person}{Wei-Ting Chen}, \bibinfo{person}{Hao-Yu Fang}, \bibinfo{person}{Cheng-Lin Hsieh}, \bibinfo{person}{Cheng-Che Tsai}, \bibinfo{person}{I Chen}, \bibinfo{person}{Jian-Jiun Ding}, \bibinfo{person}{Sy-Yen Kuo}, {et~al\mbox{.}}} \bibinfo{year}{2021}\natexlab{}.
\newblock \showarticletitle{ALL Snow Removed: Single Image Desnowing Algorithm Using Hierarchical Dual-Tree Complex Wavelet Representation and Contradict Channel Loss}. In \bibinfo{booktitle}{\emph{Proceedings of the IEEE/CVF International Conference on Computer Vision (CVPR)}}. \bibinfo{pages}{4196--4205}.
\newblock


\bibitem[Chen et~al\mbox{.}(2025)]%
        {internvl2}
\bibfield{author}{\bibinfo{person}{Zhe Chen}, \bibinfo{person}{Weiyun Wang}, \bibinfo{person}{Yue Cao}, \bibinfo{person}{Yangzhou Liu}, \bibinfo{person}{Zhangwei Gao}, \bibinfo{person}{Erfei Cui}, \bibinfo{person}{Jinguo Zhu}, \bibinfo{person}{Shenglong Ye}, \bibinfo{person}{Hao Tian}, {et~al\mbox{.}}} \bibinfo{year}{2025}\natexlab{}.
\newblock \showarticletitle{Expanding Performance Boundaries of Open-Source Multimodal Models with Model, Data, and Test-Time Scaling}.
\newblock \bibinfo{journal}{\emph{arXiv preprint arXiv:2412.05271}} (\bibinfo{year}{2025}).
\newblock


\bibitem[Chen et~al\mbox{.}(2024b)]%
        {internViT}
\bibfield{author}{\bibinfo{person}{Zhe Chen}, \bibinfo{person}{Jiannan Wu}, \bibinfo{person}{Wenhai Wang}, \bibinfo{person}{Weijie Su}, \bibinfo{person}{Guo Chen}, \bibinfo{person}{Sen Xing}, {et~al\mbox{.}}} \bibinfo{year}{2024}\natexlab{b}.
\newblock \showarticletitle{InternVL: Scaling up Vision Foundation Models and Aligning for Generic Visual-Linguistic Tasks}. In \bibinfo{booktitle}{\emph{Proceedings of the IEEE/CVF Conference on Computer Vision and Pattern Recognition (CVPR)}}. \bibinfo{pages}{24185--24198}.
\newblock


\bibitem[Chen et~al\mbox{.}(2024c)]%
        {internvl}
\bibfield{author}{\bibinfo{person}{Zhe Chen}, \bibinfo{person}{Jiannan Wu}, \bibinfo{person}{Wenhai Wang}, \bibinfo{person}{Weijie Su}, \bibinfo{person}{Guo Chen}, \bibinfo{person}{Sen Xing}, {et~al\mbox{.}}} \bibinfo{year}{2024}\natexlab{c}.
\newblock \showarticletitle{Internvl: Scaling up vision foundation models and aligning for generic visual-linguistic tasks}. In \bibinfo{booktitle}{\emph{Proceedings of the IEEE/CVF Conference on Computer Vision and Pattern Recognition (CVPR)}}. \bibinfo{pages}{24185--24198}.
\newblock


\bibitem[Chen~Wei(2018)]%
        {LOL}
\bibfield{author}{\bibinfo{person}{Wenhan Yang Jiaying~Liu Chen~Wei, Wenjing~Wang}.} \bibinfo{year}{2018}\natexlab{}.
\newblock \showarticletitle{Deep Retinex Decomposition for Low-Light Enhancement}. In \bibinfo{booktitle}{\emph{Proceedings of the British Machine Vision Conference (BMVC)}}.
\newblock


\bibitem[Damera-Venkata et~al\mbox{.}(2000)]%
        {NQM}
\bibfield{author}{\bibinfo{person}{N. Damera-Venkata}, \bibinfo{person}{T.~D. Kite}, \bibinfo{person}{W.~S. Geisler}, \bibinfo{person}{B.~L. Evans}, {and} \bibinfo{person}{A.~C. Bovik}.} \bibinfo{year}{2000}\natexlab{}.
\newblock \showarticletitle{Image quality assessment based on a degradation model}.
\newblock \bibinfo{journal}{\emph{IEEE Transactions on Image Processing (TIP)}} \bibinfo{volume}{9}, \bibinfo{number}{4} (\bibinfo{year}{2000}), \bibinfo{pages}{636--650}.
\newblock


\bibitem[Duan et~al\mbox{.}(2025)]%
        {finevq}
\bibfield{author}{\bibinfo{person}{Huiyu Duan}, \bibinfo{person}{Qiang Hu}, \bibinfo{person}{Jiarui Wang}, \bibinfo{person}{Liu Yang}, \bibinfo{person}{Zitong Xu}, \bibinfo{person}{Lu Liu}, \bibinfo{person}{Xiongkuo Min}, \bibinfo{person}{Chunlei Cai}, \bibinfo{person}{Tianxiao Ye}, \bibinfo{person}{Xiaoyun Zhang}, {and} \bibinfo{person}{Guangtao Zhai}.} \bibinfo{year}{2025}\natexlab{}.
\newblock \showarticletitle{FineVQ: Fine-Grained User Generated Content Video Quality Assessment}. In \bibinfo{booktitle}{\emph{Proceedings of the IEEE/CVF Conference on Computer Vision and Pattern Recognition (CVPR)}}.
\newblock


\bibitem[Esser et~al\mbox{.}(2024)]%
        {FLUX}
\bibfield{author}{\bibinfo{person}{Patrick Esser}, \bibinfo{person}{Sumith Kulal}, \bibinfo{person}{Andreas Blattmann}, \bibinfo{person}{Rahim Entezari}, \bibinfo{person}{Jonas M\"{u}ller}, \bibinfo{person}{Harry Saini}, {et~al\mbox{.}}} \bibinfo{year}{2024}\natexlab{}.
\newblock \showarticletitle{Scaling rectified flow transformers for high-resolution image synthesis}. In \bibinfo{booktitle}{\emph{Proceedings of the International Conference on Machine Learning (ICML)}}.
\newblock


\bibitem[Fu et~al\mbox{.}(2017)]%
        {derain}
\bibfield{author}{\bibinfo{person}{X. Fu}, \bibinfo{person}{J. Huang}, \bibinfo{person}{X. Ding}, \bibinfo{person}{Y. Liao}, {and} \bibinfo{person}{J. Paisley}.} \bibinfo{year}{2017}\natexlab{}.
\newblock \showarticletitle{Clearing the Skies: A Deep Network Architecture for Single-Image Rain Removal}.
\newblock \bibinfo{journal}{\emph{IEEE Transactions on Image Processing (TIP)}} \bibinfo{volume}{26}, \bibinfo{number}{6} (\bibinfo{year}{2017}), \bibinfo{pages}{2944--2956}.
\newblock


\bibitem[Garibi et~al\mbox{.}(2024)]%
        {renoise}
\bibfield{author}{\bibinfo{person}{Daniel Garibi}, \bibinfo{person}{Or Patashnik}, \bibinfo{person}{Andrey Voynov}, \bibinfo{person}{Hadar Averbuch-Elor}, {and} \bibinfo{person}{Daniel Cohen-Or}.} \bibinfo{year}{2024}\natexlab{}.
\newblock \showarticletitle{ReNoise: Real Image Inversion Through Iterative Noising}.
\newblock \bibinfo{journal}{\emph{arXiv preprint arXiv:2403.14602}} (\bibinfo{year}{2024}).
\newblock


\bibitem[Ghildyal and Liu(2022)]%
        {STLPIPS}
\bibfield{author}{\bibinfo{person}{Abhijay Ghildyal} {and} \bibinfo{person}{Feng Liu}.} \bibinfo{year}{2022}\natexlab{}.
\newblock \showarticletitle{Shift-Tolerant Perceptual Similarity Metric}. In \bibinfo{booktitle}{\emph{Proceedings of the European Conference on Computer Vision (ECCV)}}. \bibinfo{pages}{91–107}.
\newblock


\bibitem[Goodfellow et~al\mbox{.}(2020)]%
        {GAN}
\bibfield{author}{\bibinfo{person}{Ian Goodfellow}, \bibinfo{person}{Jean Pouget-Abadie}, \bibinfo{person}{Mehdi Mirza}, \bibinfo{person}{Bing Xu}, \bibinfo{person}{David Warde-Farley}, \bibinfo{person}{Sherjil Ozair}, \bibinfo{person}{Aaron Courville}, {and} \bibinfo{person}{Yoshua Bengio}.} \bibinfo{year}{2020}\natexlab{}.
\newblock \showarticletitle{Generative adversarial networks}.
\newblock \bibinfo{journal}{\emph{Communications of the ACM (CACM)}} \bibinfo{volume}{63}, \bibinfo{number}{11} (\bibinfo{date}{Oct.} \bibinfo{year}{2020}), \bibinfo{pages}{139–144}.
\newblock


\bibitem[Gu et~al\mbox{.}(2012)]%
        {SCSSIM}
\bibfield{author}{\bibinfo{person}{K. Gu}, \bibinfo{person}{G. Zhai}, \bibinfo{person}{X. Yang}, {and} \bibinfo{person}{W. Zhang}.} \bibinfo{year}{2012}\natexlab{}.
\newblock \showarticletitle{An improved full-reference image quality metric based on structure compensation}. In \bibinfo{booktitle}{\emph{Proceedings of the conference on Asia-Pacific Signal and Information Processing Association (APSIPA)}}. \bibinfo{pages}{1--6}.
\newblock


\bibitem[Hertz et~al\mbox{.}(2023)]%
        {DDS}
\bibfield{author}{\bibinfo{person}{Amir Hertz}, \bibinfo{person}{Kfir Aberman}, {and} \bibinfo{person}{Daniel Cohen-Or}.} \bibinfo{year}{2023}\natexlab{}.
\newblock \showarticletitle{Delta Denoising Score}. In \bibinfo{booktitle}{\emph{Proceedings of the IEEE/CVF International Conference on Computer Vision (CVPR)}}. \bibinfo{pages}{2328–2337}.
\newblock


\bibitem[Hessel et~al\mbox{.}(2021)]%
        {clipscore}
\bibfield{author}{\bibinfo{person}{Jack Hessel}, \bibinfo{person}{Ari Holtzman}, \bibinfo{person}{Maxwell Forbes}, \bibinfo{person}{Ronan~Le Bras}, {and} \bibinfo{person}{Yejin Choi}.} \bibinfo{year}{2021}\natexlab{}.
\newblock \showarticletitle{Clipscore: A Reference-Free Evaluation Metric for Image Captioning}.
\newblock \bibinfo{journal}{\emph{arXiv preprint arXiv:2104.08718}} (\bibinfo{year}{2021}).
\newblock


\bibitem[Huang et~al\mbox{.}(2024)]%
        {ESurvey}
\bibfield{author}{\bibinfo{person}{Yi Huang}, \bibinfo{person}{Jiancheng Huang}, \bibinfo{person}{Yifan Liu}, \bibinfo{person}{Mingfu Yan}, \bibinfo{person}{Jiaxi Lv}, \bibinfo{person}{Jianzhuang Liu}, {et~al\mbox{.}}} \bibinfo{year}{2024}\natexlab{}.
\newblock \showarticletitle{Diffusion Model-Based Image Editing: A Survey}.
\newblock \bibinfo{journal}{\emph{IEEE transactions on pattern analysis and machine intelligence (TPAMI)}}  \bibinfo{volume}{PP} (\bibinfo{year}{2024}).
\newblock


\bibitem[Huberman-Spiegelglas et~al\mbox{.}(2024)]%
        {DDPM}
\bibfield{author}{\bibinfo{person}{Inbar Huberman-Spiegelglas}, \bibinfo{person}{Vladimir Kulikov}, {and} \bibinfo{person}{Tomer Michaeli}.} \bibinfo{year}{2024}\natexlab{}.
\newblock \showarticletitle{An edit friendly {DDPM} noise space: Inversion and manipulations}. In \bibinfo{booktitle}{\emph{Proceedings of the IEEE/CVF Conference on Computer Vision and Pattern Recognition (CVPR)}}. \bibinfo{pages}{12469--12478}.
\newblock


\bibitem[Hui et~al\mbox{.}(2024)]%
        {HQ}
\bibfield{author}{\bibinfo{person}{Mude Hui}, \bibinfo{person}{Siwei Yang}, \bibinfo{person}{Bingchen Zhao}, \bibinfo{person}{Yichun Shi}, \bibinfo{person}{Heng Wang}, \bibinfo{person}{Peng Wang}, {et~al\mbox{.}}} \bibinfo{year}{2024}\natexlab{}.
\newblock \showarticletitle{HQ-Edit: A High-Quality Dataset for Instruction-Based Image Editing}.
\newblock \bibinfo{journal}{\emph{arXiv preprint arXiv:2404.09990}} (\bibinfo{year}{2024}).
\newblock


\bibitem[(ITU)(2012)]%
        {subject}
\bibfield{author}{\bibinfo{person}{International Telecommunication~Union (ITU)}.} \bibinfo{year}{2012}\natexlab{}.
\newblock \bibinfo{booktitle}{\emph{Methodology for the Subjective Assessment of the Quality of Television Pictures}}.
\newblock \bibinfo{type}{{T}echnical {R}eport} Rec. ITU-R BT.500-13. \bibinfo{institution}{International Telecommunication Union (ITU)}.
\newblock


\bibitem[Ju et~al\mbox{.}(2024)]%
        {PnP}
\bibfield{author}{\bibinfo{person}{Xuan Ju}, \bibinfo{person}{Ailing Zeng}, \bibinfo{person}{Yuxuan Bian}, \bibinfo{person}{Shaoteng Liu}, {and} \bibinfo{person}{Qiang Xu}.} \bibinfo{year}{2024}\natexlab{}.
\newblock \showarticletitle{PnP Inversion: Boosting Diffusion-based Editing with 3 Lines of Code}. In \bibinfo{booktitle}{\emph{Proceedings of the International Conference on Learning Representations ({ICLR})}}.
\newblock


\bibitem[Kang et~al\mbox{.}(2014)]%
        {CNN}
\bibfield{author}{\bibinfo{person}{Le Kang}, \bibinfo{person}{Peng Ye}, \bibinfo{person}{Yi Li}, {and} \bibinfo{person}{David Doermann}.} \bibinfo{year}{2014}\natexlab{}.
\newblock \showarticletitle{Convolutional Neural Networks for No-Reference Image Quality Assessment}. In \bibinfo{booktitle}{\emph{Proceedings of the IEEE/CVF Conference on Computer Vision and Pattern Recognition (CVPR)}}.
\newblock


\bibitem[Kawar et~al\mbox{.}(2023)]%
        {TedBench}
\bibfield{author}{\bibinfo{person}{Bahjat Kawar}, \bibinfo{person}{Shiran Zada}, \bibinfo{person}{Oran Lang}, \bibinfo{person}{Omer Tov}, \bibinfo{person}{Huiwen Chang}, \bibinfo{person}{Tali Dekel}, {et~al\mbox{.}}} \bibinfo{year}{2023}\natexlab{}.
\newblock \showarticletitle{Imagic: Text-Based Real Image Editing with Diffusion Models}. In \bibinfo{booktitle}{\emph{Proceedings of the IEEE/CVF Conference on Computer Vision and Pattern Recognition (CVPR)}}.
\newblock


\bibitem[Kirstain et~al\mbox{.}(2023)]%
        {pickscore}
\bibfield{author}{\bibinfo{person}{Yuval Kirstain}, \bibinfo{person}{Adam Polyak}, \bibinfo{person}{Uriel Singer}, \bibinfo{person}{Shahbuland Matiana}, \bibinfo{person}{Joe Penna}, {and} \bibinfo{person}{Omer Levy}.} \bibinfo{year}{2023}\natexlab{}.
\newblock \showarticletitle{Pick-a-Pic: An Open Dataset of User Preferences for Text-to-Image Generation}. In \bibinfo{booktitle}{\emph{Proceedings of the Advances in Neural Information Processing Systems (NeurIPS)}}. \bibinfo{pages}{36652--36663}.
\newblock


\bibitem[Koo et~al\mbox{.}(2024)]%
        {PDS}
\bibfield{author}{\bibinfo{person}{Juil Koo}, \bibinfo{person}{Chanho Park}, {and} \bibinfo{person}{Minhyuk Sung}.} \bibinfo{year}{2024}\natexlab{}.
\newblock \showarticletitle{Posterior Distillation Sampling}. In \bibinfo{booktitle}{\emph{Proceedings of the IEEE/CVF Conference on Computer Vision and Pattern Recognition (CVPR)}}. \bibinfo{pages}{13352--13361}.
\newblock


\bibitem[Kulikov et~al\mbox{.}(2024)]%
        {Flowedit}
\bibfield{author}{\bibinfo{person}{Vladimir Kulikov}, \bibinfo{person}{Matan Kleiner}, \bibinfo{person}{Inbar Huberman-Spiegelglas}, {and} \bibinfo{person}{Tomer Michaeli}.} \bibinfo{year}{2024}\natexlab{}.
\newblock \showarticletitle{FlowEdit: Inversion-Free Text-Based Editing Using Pre-Trained Flow Models}.
\newblock \bibinfo{journal}{\emph{arXiv preprint arXiv:2412.08629}} (\bibinfo{year}{2024}).
\newblock


\bibitem[Kwon et~al\mbox{.}(2023)]%
        {SLS}
\bibfield{author}{\bibinfo{person}{Mingi Kwon}, \bibinfo{person}{Jaeseok Jeong}, {and} \bibinfo{person}{Youngjung Uh}.} \bibinfo{year}{2023}\natexlab{}.
\newblock \showarticletitle{Diffusion Models Already Have a Semantic Latent Space}. In \bibinfo{booktitle}{\emph{Proceedings of the International Conference on Learning Representations (ICLR)}}.
\newblock


\bibitem[Lao et~al\mbox{.}(2022)]%
        {AHIQ}
\bibfield{author}{\bibinfo{person}{Shanshan Lao}, \bibinfo{person}{Yuan Gong}, \bibinfo{person}{Shuwei Shi}, \bibinfo{person}{Sidi Yang}, \bibinfo{person}{Tianhe Wu}, \bibinfo{person}{Jiahao Wang}, {et~al\mbox{.}}} \bibinfo{year}{2022}\natexlab{}.
\newblock \showarticletitle{Attentions Help CNNs See Better: Attention-Based Hybrid Image Quality Assessment Network}. In \bibinfo{booktitle}{\emph{Proceedings of the IEEE/CVF Conference on Computer Vision and Pattern Recognition (CVPR) Workshops}}. \bibinfo{pages}{1140--1149}.
\newblock


\bibitem[Li et~al\mbox{.}(2024a)]%
        {vqa}
\bibfield{author}{\bibinfo{person}{Baiqi Li}, \bibinfo{person}{Zhiqiu Lin}, \bibinfo{person}{Deepak Pathak}, \bibinfo{person}{Jiayao Li}, \bibinfo{person}{Yixin Fei}, \bibinfo{person}{Kewen Wu}, {et~al\mbox{.}}} \bibinfo{year}{2024}\natexlab{a}.
\newblock \showarticletitle{Evaluating and Improving Compositional Text-to-Visual Generation}. In \bibinfo{booktitle}{\emph{Proceedings of the IEEE/CVF Conference on Computer Vision and Pattern Recognition (CVPR)}}.
\newblock


\bibitem[Li et~al\mbox{.}(2024c)]%
        {llava-n}
\bibfield{author}{\bibinfo{person}{Feng Li}, \bibinfo{person}{Renrui Zhang}, \bibinfo{person}{Hao Zhang}, \bibinfo{person}{Yuanhan Zhang}, \bibinfo{person}{Bo Li}, \bibinfo{person}{Wei Li}, {et~al\mbox{.}}} \bibinfo{year}{2024}\natexlab{c}.
\newblock \showarticletitle{LLaVA-Next-Interleave: Tackling Multi-Image, Video, and 3D in Large Multimodal Models}.
\newblock \bibinfo{journal}{\emph{arXiv preprint arXiv:2407.07895}} (\bibinfo{year}{2024}).
\newblock


\bibitem[Li et~al\mbox{.}(2022)]%
        {blip}
\bibfield{author}{\bibinfo{person}{Junnan Li}, \bibinfo{person}{Dongxu Li}, \bibinfo{person}{Caiming Xiong}, {and} \bibinfo{person}{Steven Hoi}.} \bibinfo{year}{2022}\natexlab{}.
\newblock \showarticletitle{BLIP: Bootstrapping Language-Image Pre-training for Unified Vision-Language Understanding and Generation}. In \bibinfo{booktitle}{\emph{Proceedings of the International Conference on Machine Learning (ICML)}}. \bibinfo{pages}{12888--12900}.
\newblock


\bibitem[Li et~al\mbox{.}(2023a)]%
        {instructany2pix}
\bibfield{author}{\bibinfo{person}{Shufan Li}, \bibinfo{person}{Harkanwar Singh}, {and} \bibinfo{person}{Aditya Grover}.} \bibinfo{year}{2023}\natexlab{a}.
\newblock \showarticletitle{Instructany2pix: Flexible Visual Editing via Multimodal Instruction Following}.
\newblock \bibinfo{journal}{\emph{arXiv preprint arXiv:2312.06738}} (\bibinfo{year}{2023}).
\newblock


\bibitem[Li et~al\mbox{.}(2023b)]%
        {zone}
\bibfield{author}{\bibinfo{person}{Shanglin Li}, \bibinfo{person}{Bohan Zeng}, \bibinfo{person}{Yutang Feng}, \bibinfo{person}{Sicheng Gao}, \bibinfo{person}{Xuhui Liu}, \bibinfo{person}{Jiaming Liu}, {et~al\mbox{.}}} \bibinfo{year}{2023}\natexlab{b}.
\newblock \showarticletitle{Zone: Zero-Shot Instruction-Guided Local Editing}.
\newblock \bibinfo{journal}{\emph{arXiv preprint arXiv:2312.16794}} (\bibinfo{year}{2023}).
\newblock


\bibitem[Li et~al\mbox{.}(2024b)]%
        {llavascore}
\bibfield{author}{\bibinfo{person}{Yuheng Li}, \bibinfo{person}{Haotian Liu}, \bibinfo{person}{Mu Cai}, \bibinfo{person}{Yijun Li}, \bibinfo{person}{Eli Shechtman}, \bibinfo{person}{Zhe Lin}, {et~al\mbox{.}}} \bibinfo{year}{2024}\natexlab{b}.
\newblock \showarticletitle{Removing Distributional Discrepancies in Captions Improves Image-Text Alignment}.
\newblock \bibinfo{journal}{\emph{arXiv preprint arXiv:2410.00905}} (\bibinfo{year}{2024}).
\newblock


\bibitem[Liu et~al\mbox{.}(2023a)]%
        {llava}
\bibfield{author}{\bibinfo{person}{Haotian Liu}, \bibinfo{person}{Chunyuan Li}, \bibinfo{person}{Yuheng Li}, {and} \bibinfo{person}{Yong~Jae Lee}.} \bibinfo{year}{2023}\natexlab{a}.
\newblock \showarticletitle{Improved Baselines with Visual Instruction Tuning}.
\newblock \bibinfo{journal}{\emph{arXiv preprint arXiv:2310.03744}} (\bibinfo{year}{2023}).
\newblock


\bibitem[Liu et~al\mbox{.}(2023b)]%
        {Ituning}
\bibfield{author}{\bibinfo{person}{Haotian Liu}, \bibinfo{person}{Chunyuan Li}, \bibinfo{person}{Qingyang Wu}, {and} \bibinfo{person}{Yong~Jae Lee}.} \bibinfo{year}{2023}\natexlab{b}.
\newblock \showarticletitle{Visual Instruction Tuning}. In \bibinfo{booktitle}{\emph{Proceedings of the Advances in Neural Information Processing Systems (NeurIPS)}}, Vol.~\bibinfo{volume}{36}. \bibinfo{pages}{34892--34916}.
\newblock


\bibitem[Lu et~al\mbox{.}(2024)]%
        {deepseekvl}
\bibfield{author}{\bibinfo{person}{Haoyu Lu}, \bibinfo{person}{Wen Liu}, \bibinfo{person}{Bo Zhang}, \bibinfo{person}{Bingxuan Wang}, \bibinfo{person}{Kai Dong}, \bibinfo{person}{Bo Liu}, {et~al\mbox{.}}} \bibinfo{year}{2024}\natexlab{}.
\newblock \showarticletitle{DeepSeek-VL: Towards Real-World Vision-Language Understanding}.
\newblock \bibinfo{journal}{\emph{arXiv preprint arXiv:2403.05525}}  \bibinfo{volume}{6} (\bibinfo{year}{2024}).
\newblock


\bibitem[Ma et~al\mbox{.}(2024)]%
        {I2EBench}
\bibfield{author}{\bibinfo{person}{Yiwei Ma}, \bibinfo{person}{Jiayi Ji}, \bibinfo{person}{Ke Ye}, \bibinfo{person}{Weihuang Lin}, \bibinfo{person}{Yonghan Zheng}, \bibinfo{person}{Qiang Zhou}, {et~al\mbox{.}}} \bibinfo{year}{2024}\natexlab{}.
\newblock \showarticletitle{I2EBench: A Comprehensive Benchmark for Instruction-based Image Editing}. In \bibinfo{booktitle}{\emph{Proceedings of the Advances in Neural Information Processing Systems (NeurPIS)}}.
\newblock


\bibitem[Mao et~al\mbox{.}(2025)]%
        {ACE}
\bibfield{author}{\bibinfo{person}{Chaojie Mao}, \bibinfo{person}{Jingfeng Zhang}, \bibinfo{person}{Yulin Pan}, \bibinfo{person}{Zeyinzi Jiang}, \bibinfo{person}{Zhen Han}, \bibinfo{person}{Yu Liu}, {et~al\mbox{.}}} \bibinfo{year}{2025}\natexlab{}.
\newblock \showarticletitle{ACE++: Instruction-Based Image Creation and Editing via Context-Aware Content Filling}.
\newblock \bibinfo{journal}{\emph{arXiv preprint arXiv:2501.02487}} (\bibinfo{year}{2025}).
\newblock


\bibitem[Mittal et~al\mbox{.}(2011)]%
        {BRISQUE}
\bibfield{author}{\bibinfo{person}{Anish Mittal}, \bibinfo{person}{Anush~K. Moorthy}, {and} \bibinfo{person}{Alan~C. Bovik}.} \bibinfo{year}{2011}\natexlab{}.
\newblock \showarticletitle{Blind/Referenceless Image Spatial Quality Evaluator}. In \bibinfo{booktitle}{\emph{Proceedings of the Asilomar Conference on Signals, Systems and Computers (ACSSC)}}. \bibinfo{pages}{723--727}.
\newblock


\bibitem[Mittal et~al\mbox{.}(2013)]%
        {NIQE}
\bibfield{author}{\bibinfo{person}{Anish Mittal}, \bibinfo{person}{Rajiv Soundararajan}, {and} \bibinfo{person}{Alan~C. Bovik}.} \bibinfo{year}{2013}\natexlab{}.
\newblock \showarticletitle{Making a “Completely Blind” Image Quality Analyzer}.
\newblock \bibinfo{journal}{\emph{IEEE Signal Processing Letters (SPL)}} \bibinfo{volume}{20}, \bibinfo{number}{3} (\bibinfo{year}{2013}), \bibinfo{pages}{209--212}.
\newblock


\bibitem[Moorthy and Bovik(2009)]%
        {BIQI}
\bibfield{author}{\bibinfo{person}{A.~K. Moorthy} {and} \bibinfo{person}{A.~C. Bovik}.} \bibinfo{year}{2009}\natexlab{}.
\newblock \showarticletitle{A Modular Framework for Constructing Blind Universal Quality Indices}.
\newblock \bibinfo{journal}{\emph{IEEE Signal Processing Letters (SPL)}} (\bibinfo{year}{2009}).
\newblock


\bibitem[Moorthy and Bovik(2011)]%
        {DIIVINE}
\bibfield{author}{\bibinfo{person}{Anush~Krishna Moorthy} {and} \bibinfo{person}{Alan~Conrad Bovik}.} \bibinfo{year}{2011}\natexlab{}.
\newblock \showarticletitle{Blind Image Quality Assessment: From Natural Scene Statistics to Perceptual Quality}.
\newblock \bibinfo{journal}{\emph{IEEE Transactions on Image Processing (TIP)}} \bibinfo{volume}{20}, \bibinfo{number}{12} (\bibinfo{year}{2011}), \bibinfo{pages}{3350--3364}.
\newblock


\bibitem[Nah et~al\mbox{.}(2017)]%
        {GoPro}
\bibfield{author}{\bibinfo{person}{Seungjun Nah}, \bibinfo{person}{Tae~Hyun Kim}, {and} \bibinfo{person}{Kyoung~Mu Lee}.} \bibinfo{year}{2017}\natexlab{}.
\newblock \showarticletitle{Deep Multi-Scale Convolutional Neural Network for Dynamic Scene Deblurring}. In \bibinfo{booktitle}{\emph{Proceedings of the IEEE/CVF Conference on Computer Vision and Pattern Recognition (CVPR)}}.
\newblock


\bibitem[Nam et~al\mbox{.}(2024)]%
        {CDS}
\bibfield{author}{\bibinfo{person}{Hyelin Nam}, \bibinfo{person}{Gihyun Kwon}, \bibinfo{person}{Geon~Yeong Park}, {and} \bibinfo{person}{Jong~Chul Ye}.} \bibinfo{year}{2024}\natexlab{}.
\newblock \showarticletitle{Contrastive Denoising Score for Text-guided Latent Diffusion Image Editing}. In \bibinfo{booktitle}{\emph{Proceedings of the IEEE/CVF Conference on Computer Vision and Pattern Recognition (CVPR)}}. \bibinfo{pages}{9192--9201}.
\newblock


\bibitem[OpenAI et~al\mbox{.}(2024)]%
        {gpt4}
\bibfield{author}{\bibinfo{person}{OpenAI}, \bibinfo{person}{Josh Achiam}, \bibinfo{person}{Steven Adler}, \bibinfo{person}{Sandhini Agarwal}, \bibinfo{person}{Lama Ahmad}, \bibinfo{person}{Ilge Akkaya}, \bibinfo{person}{Florencia~Leoni Aleman}, {et~al\mbox{.}}} \bibinfo{year}{2024}\natexlab{}.
\newblock \showarticletitle{GPT-4 Technical Report}.
\newblock \bibinfo{journal}{\emph{arXiv preprint arXiv:2303.08774}} (\bibinfo{year}{2024}).
\newblock


\bibitem[Podell et~al\mbox{.}(2023)]%
        {sdxl}
\bibfield{author}{\bibinfo{person}{Dustin Podell}, \bibinfo{person}{Zion English}, \bibinfo{person}{Kyle Lacey}, \bibinfo{person}{Andreas Blattmann}, \bibinfo{person}{Tim Dockhorn}, \bibinfo{person}{Jonas Müller}, {et~al\mbox{.}}} \bibinfo{year}{2023}\natexlab{}.
\newblock \showarticletitle{Sdxl: Improving Latent Diffusion Models for High-Resolution Image Synthesis}.
\newblock \bibinfo{journal}{\emph{arXiv preprint arXiv:2307.01952}} (\bibinfo{year}{2023}).
\newblock


\bibitem[Radford et~al\mbox{.}(2021)]%
        {ViT}
\bibfield{author}{\bibinfo{person}{Alec Radford}, \bibinfo{person}{Jong~Wook Kim}, \bibinfo{person}{Chris Hallacy}, \bibinfo{person}{A. Ramesh}, \bibinfo{person}{Gabriel Goh}, \bibinfo{person}{Sandhini Agarwal}, \bibinfo{person}{Girish Sastry}, \bibinfo{person}{Amanda Askell}, \bibinfo{person}{Pamela Mishkin}, \bibinfo{person}{Jack Clark}, \bibinfo{person}{Gretchen Krueger}, {and} \bibinfo{person}{Ilya Sutskever}.} \bibinfo{year}{2021}\natexlab{}.
\newblock \showarticletitle{Learning Transferable Visual Models From Natural Language Supervision}. In \bibinfo{booktitle}{\emph{Proceedings of the International Conference on Machine Learning (ICML)}}.
\newblock


\bibitem[Rombach et~al\mbox{.}(2022)]%
        {SD}
\bibfield{author}{\bibinfo{person}{Robin Rombach}, \bibinfo{person}{Andreas Blattmann}, \bibinfo{person}{Dominik Lorenz}, \bibinfo{person}{Patrick Esser}, {and} \bibinfo{person}{Bj{\"o}rn Ommer}.} \bibinfo{year}{2022}\natexlab{}.
\newblock \showarticletitle{High-resolution image synthesis with latent diffusion models}. In \bibinfo{booktitle}{\emph{Proceedings of the IEEE/CVF Conference on Computer Vision and Pattern Recognition (CVPR)}}. \bibinfo{pages}{10684--10695}.
\newblock


\bibitem[Saad et~al\mbox{.}(2012)]%
        {BLII}
\bibfield{author}{\bibinfo{person}{Michele~A. Saad}, \bibinfo{person}{Alan~C. Bovik}, {and} \bibinfo{person}{Christophe Charrier}.} \bibinfo{year}{2012}\natexlab{}.
\newblock \showarticletitle{Blind Image Quality Assessment: A Natural Scene Statistics Approach in the DCT Domain}.
\newblock \bibinfo{journal}{\emph{IEEE Transactions on Image Processing (TIP)}} \bibinfo{volume}{21}, \bibinfo{number}{8} (\bibinfo{year}{2012}), \bibinfo{pages}{3339--3352}.
\newblock


\bibitem[Sheikh and Bovik(2006)]%
        {VIF}
\bibfield{author}{\bibinfo{person}{H.~R. Sheikh} {and} \bibinfo{person}{A.~C. Bovik}.} \bibinfo{year}{2006}\natexlab{}.
\newblock \showarticletitle{Image information and visual quality}.
\newblock \bibinfo{journal}{\emph{IEEE Transactions on Image Processing (TIP)}} \bibinfo{volume}{15}, \bibinfo{number}{2} (\bibinfo{year}{2006}), \bibinfo{pages}{430--444}.
\newblock


\bibitem[Sheikh et~al\mbox{.}(2005)]%
        {IFC}
\bibfield{author}{\bibinfo{person}{H.~R. Sheikh}, \bibinfo{person}{A.~C. Bovik}, {and} \bibinfo{person}{G.~De Veciana}.} \bibinfo{year}{2005}\natexlab{}.
\newblock \showarticletitle{An information fidelity criterion for image quality assessment using natural scene statistics}.
\newblock \bibinfo{journal}{\emph{IEEE Transactions on Image Processing (TIP)}} \bibinfo{volume}{14}, \bibinfo{number}{12} (\bibinfo{year}{2005}), \bibinfo{pages}{2117--2128}.
\newblock


\bibitem[Sheynin et~al\mbox{.}(2024)]%
        {EmuEdit}
\bibfield{author}{\bibinfo{person}{Shelly Sheynin}, \bibinfo{person}{Adam Polyak}, \bibinfo{person}{Uriel Singer}, \bibinfo{person}{Yuval Kirstain}, \bibinfo{person}{Amit Zohar}, \bibinfo{person}{Oron Ashual}, \bibinfo{person}{Devi Parikh}, {and} \bibinfo{person}{Yaniv Taigman}.} \bibinfo{year}{2024}\natexlab{}.
\newblock \showarticletitle{Emu Edit: Precise Image Editing via Recognition and Generation Tasks}. In \bibinfo{booktitle}{\emph{Proceedings of the IEEE/CVF Conference on Computer Vision and Pattern Recognition (CVPR)}}. \bibinfo{pages}{8871--8879}.
\newblock


\bibitem[Su et~al\mbox{.}(2020)]%
        {Hyper}
\bibfield{author}{\bibinfo{person}{S. Su}, \bibinfo{person}{Q. Yan}, \bibinfo{person}{Y. Zhu}, \bibinfo{person}{C. Zhang}, \bibinfo{person}{X. Ge}, \bibinfo{person}{J. Sun}, {et~al\mbox{.}}} \bibinfo{year}{2020}\natexlab{}.
\newblock \showarticletitle{Blindly assess image quality in the wild guided by a self-adaptive hyper network}. In \bibinfo{booktitle}{\emph{Proceedings of the IEEE/CVF Conference on Computer Vision and Pattern Recognition (CVPR)}}.
\newblock


\bibitem[Sun et~al\mbox{.}(2025)]%
        {IEBench}
\bibfield{author}{\bibinfo{person}{Shangkun Sun}, \bibinfo{person}{Bowen Qu}, \bibinfo{person}{Xiaoyu Liang}, \bibinfo{person}{Songlin Fan}, {and} \bibinfo{person}{Wei Gao}.} \bibinfo{year}{2025}\natexlab{}.
\newblock \showarticletitle{IE-Bench: Advancing the Measurement of Text-Driven Image Editing for Human Perception Alignment}.
\newblock \bibinfo{journal}{\emph{arXiv preprint arXiv:2501.09927}} (\bibinfo{year}{2025}).
\newblock


\bibitem[Talebi and Milanfar(2018)]%
        {NIMA}
\bibfield{author}{\bibinfo{person}{H. Talebi} {and} \bibinfo{person}{P. Milanfar}.} \bibinfo{year}{2018}\natexlab{}.
\newblock \showarticletitle{NIMA: Neural image assessment}.
\newblock \bibinfo{journal}{\emph{IEEE Transactions on Image Processing (TIP)}} \bibinfo{volume}{27}, \bibinfo{number}{8} (\bibinfo{year}{2018}), \bibinfo{pages}{3998--4011}.
\newblock


\bibitem[Tang et~al\mbox{.}(2023)]%
        {DIFT}
\bibfield{author}{\bibinfo{person}{Luming Tang}, \bibinfo{person}{Menglin Jia}, \bibinfo{person}{Qianqian Wang}, \bibinfo{person}{Cheng~Perng Phoo}, {and} \bibinfo{person}{Bharath Hariharan}.} \bibinfo{year}{2023}\natexlab{}.
\newblock \showarticletitle{Emergent Correspondence from Image Diffusion}. In \bibinfo{booktitle}{\emph{Proceedings of the Conference on Neural Information Processing Systems (NeurIPS)}}.
\newblock


\bibitem[Wallace et~al\mbox{.}(2022)]%
        {EDICT}
\bibfield{author}{\bibinfo{person}{Bram Wallace}, \bibinfo{person}{Akash Gokul}, {and} \bibinfo{person}{Nikhil Naik}.} \bibinfo{year}{2022}\natexlab{}.
\newblock \showarticletitle{EDICT: Exact Diffusion Inversion via Coupled Transformations}.
\newblock \bibinfo{journal}{\emph{arXiv preprint arXiv:2211.12446}} (\bibinfo{year}{2022}).
\newblock


\bibitem[Wang et~al\mbox{.}(2023a)]%
        {CLIPIQA}
\bibfield{author}{\bibinfo{person}{Jianyi Wang}, \bibinfo{person}{Kelvin~C.K. Chan}, {and} \bibinfo{person}{Chen~Change Loy}.} \bibinfo{year}{2023}\natexlab{a}.
\newblock \showarticletitle{Exploring CLIP for assessing the look and feel of images}. In \bibinfo{booktitle}{\emph{Proceedings of the Conference on Association for the Advancement of Artificial Intelligence (AAAI)}}.
\newblock


\bibitem[Wang et~al\mbox{.}(2023b)]%
        {aigciqa2023}
\bibfield{author}{\bibinfo{person}{Jiarui Wang}, \bibinfo{person}{Huiyu Duan}, \bibinfo{person}{Jing Liu}, \bibinfo{person}{Shi Chen}, \bibinfo{person}{Xiongkuo Min}, {and} \bibinfo{person}{Guangtao Zhai}.} \bibinfo{year}{2023}\natexlab{b}.
\newblock \showarticletitle{AIGCIQA2023: A Large-Scale Image Quality Assessment Database for AI Generated Images: From the Perspectives of Quality, Authenticity and Correspondence}. In \bibinfo{booktitle}{\emph{Proceedings of the CAAI International Conference on Artificial Intelligence (CICAI)}}. \bibinfo{pages}{46–57}.
\newblock


\bibitem[Wang et~al\mbox{.}(2025a)]%
        {IQAinstruction}
\bibfield{author}{\bibinfo{person}{Jiarui Wang}, \bibinfo{person}{Huiyu Duan}, \bibinfo{person}{Guangtao Zhai}, {and} \bibinfo{person}{Xiongkuo Min}.} \bibinfo{year}{2025}\natexlab{a}.
\newblock \showarticletitle{Quality Assessment for AI Generated Images with Instruction Tuning}.
\newblock \bibinfo{journal}{\emph{IEEE Transactions on Multimedia (TMM)}} (\bibinfo{year}{2025}).
\newblock


\bibitem[Wang et~al\mbox{.}(2025b)]%
        {aigv}
\bibfield{author}{\bibinfo{person}{Jiarui Wang}, \bibinfo{person}{Huiyu Duan}, \bibinfo{person}{Guangtao Zhai}, \bibinfo{person}{Juntong Wang}, {and} \bibinfo{person}{Xiongkuo Min}.} \bibinfo{year}{2025}\natexlab{b}.
\newblock \showarticletitle{AIGV-Assessor: Benchmarking and Evaluating the Perceptual Quality of Text-to-Video Generation with LMM}. In \bibinfo{booktitle}{\emph{Proceedings of the IEEE/CVF Conference on Computer Vision and Pattern Recognition (CVPR)}}.
\newblock


\bibitem[Wang et~al\mbox{.}(2018)]%
        {ISTD}
\bibfield{author}{\bibinfo{person}{Jifeng Wang}, \bibinfo{person}{Xiang Li}, {and} \bibinfo{person}{Jian Yang}.} \bibinfo{year}{2018}\natexlab{}.
\newblock \showarticletitle{Stacked Conditional Generative Adversarial Networks for Jointly Learning Shadow Detection and Shadow Removal}. In \bibinfo{booktitle}{\emph{Proceedings of the IEEE/CVF International Conference on Computer Vision (CVPR)}}.
\newblock


\bibitem[Wang et~al\mbox{.}(2024b)]%
        {RFSE}
\bibfield{author}{\bibinfo{person}{Jiangshan Wang}, \bibinfo{person}{Junfu Pu}, \bibinfo{person}{Zhongang Qi}, \bibinfo{person}{Jiayi Guo}, \bibinfo{person}{Yue Ma}, \bibinfo{person}{Nisha Huang}, \bibinfo{person}{Yuxin Chen}, \bibinfo{person}{Xiu Li}, {and} \bibinfo{person}{Ying Shan}.} \bibinfo{year}{2024}\natexlab{b}.
\newblock \showarticletitle{Taming Rectified Flow for Inversion and Editing}.
\newblock \bibinfo{journal}{\emph{arXiv preprint arXiv:2411.04746}} (\bibinfo{year}{2024}).
\newblock


\bibitem[Wang et~al\mbox{.}(2024a)]%
        {qwenvl}
\bibfield{author}{\bibinfo{person}{Peng Wang}, \bibinfo{person}{Shuai Bai}, \bibinfo{person}{Sinan Tan}, \bibinfo{person}{Shijie Wang}, \bibinfo{person}{Zhihao Fan}, \bibinfo{person}{Jinze Bai}, {et~al\mbox{.}}} \bibinfo{year}{2024}\natexlab{a}.
\newblock \showarticletitle{Qwen2-VL: Enhancing Vision-Language Model’s Perception of the World at Any Resolution}.
\newblock \bibinfo{journal}{\emph{arXiv preprint arXiv:2409.12191}} (\bibinfo{year}{2024}).
\newblock


\bibitem[Wang et~al\mbox{.}(2024c)]%
        {LMMIQA}
\bibfield{author}{\bibinfo{person}{Puyi Wang}, \bibinfo{person}{Wei Sun}, \bibinfo{person}{Zicheng Zhang}, \bibinfo{person}{Jun Jia}, \bibinfo{person}{Yanwei Jiang}, \bibinfo{person}{Zhichao Zhang}, {et~al\mbox{.}}} \bibinfo{year}{2024}\natexlab{c}.
\newblock \showarticletitle{Large Multi-modality Model Assisted AI-Generated Image Quality Assessment}. In \bibinfo{booktitle}{\emph{Proceedings of the ACM International Conference on Multimedia (ACM MM)}}. \bibinfo{pages}{7803–7812}.
\newblock


\bibitem[Wang et~al\mbox{.}(2004)]%
        {SSIM}
\bibfield{author}{\bibinfo{person}{Z. Wang}, \bibinfo{person}{A.~C. Bovik}, \bibinfo{person}{H.~R. Sheikh}, {and} \bibinfo{person}{E.~P. Simoncelli}.} \bibinfo{year}{2004}\natexlab{}.
\newblock \showarticletitle{Image quality assessment: from error visibility to structural similarity}.
\newblock \bibinfo{journal}{\emph{IEEE Transactions on Image Processing (TIP)}} \bibinfo{volume}{13}, \bibinfo{number}{4} (\bibinfo{year}{2004}), \bibinfo{pages}{600--612}.
\newblock


\bibitem[Wang et~al\mbox{.}(2003)]%
        {MSSIM}
\bibfield{author}{\bibinfo{person}{Z. Wang}, \bibinfo{person}{E.~P. Simoncelli}, {and} \bibinfo{person}{A.~C. Bovik}.} \bibinfo{year}{2003}\natexlab{}.
\newblock \showarticletitle{Multiscale structural similarity for image quality assessment}. In \bibinfo{booktitle}{\emph{Proceedings of the Asilomar Conference on Signals, Systems \& Computers (ACSSC)}}, Vol.~\bibinfo{volume}{2}. \bibinfo{pages}{1398--1402}.
\newblock


\bibitem[Wu et~al\mbox{.}(2023b)]%
        {qalign}
\bibfield{author}{\bibinfo{person}{Haoning Wu}, \bibinfo{person}{Zicheng Zhang}, \bibinfo{person}{Weixia Zhang}, \bibinfo{person}{Chaofeng Chen}, \bibinfo{person}{Chunyi Li}, \bibinfo{person}{Liang Liao}, {et~al\mbox{.}}} \bibinfo{year}{2023}\natexlab{b}.
\newblock \showarticletitle{Q-Align: Teaching LMMs for Visual Scoring via Discrete Text-Defined Levels}.
\newblock \bibinfo{journal}{\emph{arXiv preprint arXiv:2312.17090}} (\bibinfo{year}{2023}).
\newblock


\bibitem[Wu et~al\mbox{.}(2023a)]%
        {HPS}
\bibfield{author}{\bibinfo{person}{Xiaoshi Wu}, \bibinfo{person}{Keqiang Sun}, \bibinfo{person}{Feng Zhu}, \bibinfo{person}{Rui Zhao}, {and} \bibinfo{person}{Hongsheng Li}.} \bibinfo{year}{2023}\natexlab{a}.
\newblock \showarticletitle{Human Preference Score: Better Aligning Text-to-Image Models with Human Preference}. In \bibinfo{booktitle}{\emph{IEEE/CVF Conference on Computer Vision and Pattern Recognition (CVPR)}}.
\newblock


\bibitem[Wu et~al\mbox{.}(2024)]%
        {deepseekvl2}
\bibfield{author}{\bibinfo{person}{Zhiyu Wu}, \bibinfo{person}{Xiaokang Chen}, \bibinfo{person}{Zizheng Pan}, \bibinfo{person}{Xingchao Liu}, \bibinfo{person}{Wen Liu}, \bibinfo{person}{Damai Dai}, {et~al\mbox{.}}} \bibinfo{year}{2024}\natexlab{}.
\newblock \showarticletitle{DeepSeek-VL2: Mixture-of-Experts Vision-Language Models for Advanced Multimodal Understanding}.
\newblock \bibinfo{journal}{\emph{arXiv preprint arXiv:2412.10302}} (\bibinfo{year}{2024}).
\newblock


\bibitem[Xu et~al\mbox{.}(2023)]%
        {imagereward}
\bibfield{author}{\bibinfo{person}{Jiazheng Xu}, \bibinfo{person}{Xiao Liu}, \bibinfo{person}{Yuchen Wu}, \bibinfo{person}{Yuxuan Tong}, \bibinfo{person}{Qinkai Li}, \bibinfo{person}{Ming Ding}, {et~al\mbox{.}}} \bibinfo{year}{2023}\natexlab{}.
\newblock \showarticletitle{ImageReward: learning and evaluating human preferences for text-to-image generation}. In \bibinfo{booktitle}{\emph{Proceedings of the International Conference on Neural Information Processing Systems (NeurIPS)}}. \bibinfo{pages}{15903--15935}.
\newblock


\bibitem[Xu et~al\mbox{.}(2024)]%
        {InfEdit}
\bibfield{author}{\bibinfo{person}{Sihan Xu}, \bibinfo{person}{Yidong Huang}, \bibinfo{person}{Jiayi Pan}, \bibinfo{person}{Ziqiao Ma}, {and} \bibinfo{person}{Joyce Chai}.} \bibinfo{year}{2024}\natexlab{}.
\newblock \showarticletitle{Inversion-Free Image Editing with Natural Language}. In \bibinfo{booktitle}{\emph{Proceedings of the IEEE/CVF Conference on Computer Vision and Pattern Recognition (CVPR)}}. \bibinfo{pages}{9192--9201}.
\newblock


\bibitem[Xue et~al\mbox{.}(2013)]%
        {GMSD}
\bibfield{author}{\bibinfo{person}{W. Xue}, \bibinfo{person}{L. Zhang}, \bibinfo{person}{X. Mou}, {and} \bibinfo{person}{A.~C. Bovik}.} \bibinfo{year}{2013}\natexlab{}.
\newblock \showarticletitle{Gradient magnitude similarity deviation: A highly efficient perceptual image quality index}.
\newblock \bibinfo{journal}{\emph{IEEE Transactions on Image Processing (TIP)}} \bibinfo{volume}{23}, \bibinfo{number}{2} (\bibinfo{year}{2013}), \bibinfo{pages}{684--695}.
\newblock


\bibitem[Yang et~al\mbox{.}(2022)]%
        {MANIQA}
\bibfield{author}{\bibinfo{person}{Sidi Yang}, \bibinfo{person}{Tianhe Wu}, \bibinfo{person}{Shuwei Shi}, \bibinfo{person}{Shanshan Lao}, \bibinfo{person}{Yuan Gong}, \bibinfo{person}{Mingdeng Cao}, {et~al\mbox{.}}} \bibinfo{year}{2022}\natexlab{}.
\newblock \showarticletitle{MANIQA: Multi-dimension Attention Network for No-Reference Image Quality Assessment}. In \bibinfo{booktitle}{\emph{Proceedings of the IEEE/CVF Conference on Computer Vision and Pattern Recognition (CVPR)}}. \bibinfo{pages}{1191--1200}.
\newblock


\bibitem[Yao et~al\mbox{.}(2024)]%
        {minicpm}
\bibfield{author}{\bibinfo{person}{Yuan Yao}, \bibinfo{person}{Tianyu Yu}, \bibinfo{person}{Ao Zhang}, \bibinfo{person}{Chongyi Wang}, \bibinfo{person}{Junbo Cui}, \bibinfo{person}{Hongji Zhu}, {et~al\mbox{.}}} \bibinfo{year}{2024}\natexlab{}.
\newblock \showarticletitle{MiniCPM-V: A GPT-4V Level MLLM on Your Phone}.
\newblock \bibinfo{journal}{\emph{arXiv preprint arXiv:2408.01800}} (\bibinfo{year}{2024}).
\newblock


\bibitem[Ye et~al\mbox{.}(2024)]%
        {mplug}
\bibfield{author}{\bibinfo{person}{Jiabo Ye}, \bibinfo{person}{Haiyang Xu}, \bibinfo{person}{Haowei Liu}, \bibinfo{person}{Anwen Hu}, \bibinfo{person}{Ming Yan}, \bibinfo{person}{Qi Qian}, {et~al\mbox{.}}} \bibinfo{year}{2024}\natexlab{}.
\newblock \showarticletitle{mPLUG-Owl3: Towards Long Image-Sequence Understanding in Multimodal Large Language Models}. In \bibinfo{booktitle}{\emph{Proceedings of the International Conference on Learning Representations (ICLR)}}.
\newblock


\bibitem[Yin et~al\mbox{.}(2022)]%
        {CVRKD}
\bibfield{author}{\bibinfo{person}{Guanghao Yin}, \bibinfo{person}{Wei Wang}, \bibinfo{person}{Zehuan Yuan}, \bibinfo{person}{Chuchu Han}, \bibinfo{person}{Wei Ji}, \bibinfo{person}{Shouqian Sun}, {et~al\mbox{.}}} \bibinfo{year}{2022}\natexlab{}.
\newblock \showarticletitle{Content-variant reference image quality assessment via knowledge distillation}. In \bibinfo{booktitle}{\emph{Proceedings of the Conference on Association for the Advancement of Artificial Intelligence (AAAI)}}, Vol.~\bibinfo{volume}{36}. \bibinfo{pages}{3134--3142}.
\newblock


\bibitem[Zhang et~al\mbox{.}(2023c)]%
        {Magicbrush}
\bibfield{author}{\bibinfo{person}{Kai Zhang}, \bibinfo{person}{Lingbo Mo}, \bibinfo{person}{Wenhu Chen}, \bibinfo{person}{Huan Sun}, {and} \bibinfo{person}{Yu Su}.} \bibinfo{year}{2023}\natexlab{c}.
\newblock \showarticletitle{Magicbrush: A manually annotated dataset for instruction-guided image editing}. In \bibinfo{booktitle}{\emph{Proceedings of the Advances in Neural Information Processing Systems (NeurPIS)}}.
\newblock


\bibitem[Zhang et~al\mbox{.}(2011)]%
        {FSIM}
\bibfield{author}{\bibinfo{person}{L. Zhang}, \bibinfo{person}{L. Zhang}, \bibinfo{person}{X. Mou}, {and} \bibinfo{person}{D. Zhang}.} \bibinfo{year}{2011}\natexlab{}.
\newblock \showarticletitle{FSIM: A feature similarity index for image quality assessment}.
\newblock \bibinfo{journal}{\emph{IEEE Transactions on Image Processing (TIP)}} \bibinfo{volume}{20}, \bibinfo{number}{8} (\bibinfo{year}{2011}), \bibinfo{pages}{2378--2386}.
\newblock


\bibitem[Zhang et~al\mbox{.}(2023b)]%
        {internlmxcomposer}
\bibfield{author}{\bibinfo{person}{Pan Zhang}, \bibinfo{person}{Xiaoyi Dong}, \bibinfo{person}{Bin Wang}, \bibinfo{person}{Yuhang Cao}, \bibinfo{person}{Chao Xu}, \bibinfo{person}{Linke Ouyang}, {et~al\mbox{.}}} \bibinfo{year}{2023}\natexlab{b}.
\newblock \showarticletitle{InternLM-XComposer: A Vision-Language Large Model for Advanced Text-Image Comprehension and Composition}.
\newblock \bibinfo{journal}{\emph{arXiv preprint arXiv:2309.15112}} (\bibinfo{year}{2023}).
\newblock


\bibitem[Zhang et~al\mbox{.}(2023a)]%
        {adalora}
\bibfield{author}{\bibinfo{person}{Qingru Zhang}, \bibinfo{person}{Minshuo Chen}, \bibinfo{person}{Alexander Bukharin}, \bibinfo{person}{Pengcheng He}, \bibinfo{person}{Yu Cheng}, \bibinfo{person}{Weizhu Chen}, {et~al\mbox{.}}} \bibinfo{year}{2023}\natexlab{a}.
\newblock \showarticletitle{Adaptive Budget Allocation for Parameter-Efficient Fine-Tuning}. In \bibinfo{booktitle}{\emph{Proceedings of the International Conference on Learning Representations (ICLR)}}.
\newblock


\bibitem[Zhang et~al\mbox{.}(2018)]%
        {LPIPS}
\bibfield{author}{\bibinfo{person}{R. Zhang}, \bibinfo{person}{P. Isola}, \bibinfo{person}{A.~A. Efros}, \bibinfo{person}{E. Shechtman}, {and} \bibinfo{person}{O. Wang}.} \bibinfo{year}{2018}\natexlab{}.
\newblock \showarticletitle{The unreasonable effectiveness of deep features as a perceptual metric}. In \bibinfo{booktitle}{\emph{Proceedings of the IEEE/CVF Conference on Computer Vision and Pattern Recognition (CVPR)}}.
\newblock


\bibitem[Zhang et~al\mbox{.}(2020)]%
        {DBCNN}
\bibfield{author}{\bibinfo{person}{W. Zhang}, \bibinfo{person}{K. Ma}, \bibinfo{person}{J. Yan}, \bibinfo{person}{D. Deng}, {and} \bibinfo{person}{Z. Wang}.} \bibinfo{year}{2020}\natexlab{}.
\newblock \showarticletitle{Blind image quality assessment using a deep bilinear convolutional neural network}.
\newblock \bibinfo{journal}{\emph{IEEE Transactions on Circuits and Systems for Video Technology (TCSVT)}} \bibinfo{volume}{30}, \bibinfo{number}{1} (\bibinfo{year}{2020}), \bibinfo{pages}{36--47}.
\newblock


\bibitem[Zhao et~al\mbox{.}(2019)]%
        {exBeDDE}
\bibfield{author}{\bibinfo{person}{S. Zhao}, \bibinfo{person}{L. Zhang}, {et~al\mbox{.}}} \bibinfo{year}{2019}\natexlab{}.
\newblock \showarticletitle{Evaluation of Defogging: A Real-World Benchmark Dataset, a New Criterion and Baselines}. In \bibinfo{booktitle}{\emph{Proceedings of the IEEE International Conference on Multimedia and Expo (ICME)}}. \bibinfo{pages}{1840--1845}.
\newblock


\end{thebibliography}


\appendix
\end{document}